\documentclass[sigconf]{acmart}

\AtBeginDocument{%
  }

\setcopyright{acmlicensed}
\copyrightyear{2018}
\acmYear{2018}
\acmDOI{XXXXXXX.XXXXXXX}

\acmConference[Conference acronym 'XX]{Make sure to enter the correct
  conference title from your rights confirmation emai}{June 03--05,
  2018}{Woodstock, NY}
\acmISBN{978-1-4503-XXXX-X/18/06}

\usepackage{subcaption}
\usepackage{multirow}
\usepackage{amsmath}
\usepackage{amsthm}
\newtheorem{definition}{Definition}
\newtheorem{proposition}{Proposition}
\usepackage{hyperref}

\begin{document}

\title{ScaDyG: A New Paradigm for Large-scale Dynamic Graph Learning }


\author{Xiang Wu, Xunkai Li, Rong-Hua Li, Kangfei Zhao, Guoren Wang}
\renewcommand{\shortauthors}{Xiang Wu et al.}

\begin{abstract}
    Dynamic graphs (DGs), which capture time-evolving relationships between graph entities, have widespread real-world applications.
    To efficiently encode DGs for downstream tasks, most dynamic graph neural networks follow the traditional message-passing mechanism and extend it with time-based techniques.
    Despite their effectiveness, the growth of historical interactions introduces significant scalability issues, particularly in industry scenarios.
    To address this limitation, we propose ScaDyG, with the core idea of designing a time-aware scalable learning paradigm as follows: 
    1) Time-aware Topology Reformulation: 
    ScaDyG first segments historical interactions into time steps (intra and inter) based on dynamic modeling, enabling weight-free and time-aware graph propagation within pre-processing.
    2) Dynamic Temporal Encoding: 
    To further achieve fine-grained graph propagation within time steps, ScaDyG integrates temporal encoding through a combination of exponential functions in a scalable manner.
    3) Hypernetwork-driven Message Aggregation:
    After obtaining the propagated features (i.e., messages), ScaDyG utilizes hypernetwork to analyze historical dependencies, implementing node-wise representation by an adaptive temporal fusion.
    Extensive experiments on 12 datasets demonstrate that ScaDyG performs comparably well or even outperforms other SOTA methods in both node and link-level downstream tasks, with fewer learnable parameters and higher efficiency. %
\end{abstract}

%%现有方法大多通过，在大规模图上历史交互的爆炸性增加为现有方法提出了重大挑战。在这个工作中，我们提出一种可扩展的动态图表示学习范式。
%%Most existing dynamic graph neural networks utilize a sampling-aggregation
%paradigm, where a subset of nodes are first sampled from the historical neighbors of a node, followed by the temporal aggregationof these sampled neighbors to obtain the current representation ofthe node.

%\keywords{Dynamic graph learning}

\maketitle

\section{Introduction}
    Recently, dynamic graphs (DGs) are widely used in social analysis~\cite{panzarasa2009:patterns, peng2020:spatial}, recommendation~\cite{zhang2022:dynamic, tang2023:dynamic}, and financial management~\cite{pareja2020:evolvegcn, xiang2022:temporal}. 
    As a type of high-order relational data, DGs capture the time-evolving relationships between graph entities, providing insights into the dynamic nature of complex systems.
    To encode DGs, dynamic graph neural networks (DGNNs) are designed to model temporal topology and integrate time-evolving node profiles.

    To provide a clear presentation, we first conduct a systematic review of the most prevalent DGNNs and propose the following taxonomy:
    (1) Discrete-based methods~\cite{zhu2023:wingnn,you2022:roland,sankar2020:dysat,pareja2020:evolvegcn} divide a dynamic graph into a sequence of snapshots, with each snapshot treated as an individual static graph.
    Based on this, these approaches model each snapshot by time-independent GNN (e.g., GAT~\cite{velivckovic2017:GAT}), while the temporal dependencies between snapshots are captured with sequence models such as RNNs~\cite{pareja2020:evolvegcn,you2022:roland}. 
    Despite their simplicity and intuitiveness, they overlook the fine-grained dynamic information within each snapshot.
    (2) Continuous-based methods \cite{luo2022neighborhood,yu2023towards,zuo2018:embedding,lu2019:temporal,wen2022:trend,xu2020:inductive} emphasize time-based dynamics and focus on interaction granularity, enabling nuanced modeling of temporal dependencies. 
    Specifically, they sample informative historical neighbors and aggregate nodes with learnable aggregators, achieving SOTA performance due to their fine temporal resolution~\cite{yu2023towards}. 
    However, these methods struggle with large-scale DGs because the algorithm complexity of sampling and aggregation scales linearly with the number of historical neighbors~\cite{li2023:zebra,li2023orca}.
    Moreover, they frequently compute node embeddings at each interaction timestamp, leading to redundant calculations. 
    Consequently, most DGNNs are constrained in their scalability to handle large-scale DGs.

    During our investigation, we found that recent advancements have introduced techniques to improve the scalability of DGNNs. 
    Specifically, they improve sampling efficiency~\cite{zhou2022:tgl,li2023:zebra} or avoid redundant calculations~\cite{li2023orca}.
    Despite their effectiveness, they still rely on the sampling-aggregation framework, which is constrained by sampling quality, especially when sample sizes are much smaller than the historical neighbors in large-scale scenarios. 
    In such cases, preserving all interactions is necessary to achieve satisfactory performance.
    However, the potentially thousands of historical neighbors in million-level DGs lead to unaffordable computational costs.

    To break the above limitations, we draw inspiration from decoupled scalable GNNs in time-independent graphs~\cite{wu2019:simplifying,frasca:2020sign,zhang2022:graph,chen2020scalable,feng2022grand+,zhu2020:simple}, which reduce algorithm complexity by separating feature propagation from learnable transformations. 
    Specifically, by designing advanced propagation operators and learnable message aggregators, these methods have achieved SOTA performance across various tasks~\cite{li2024:lightdic,li2024_atp,yang2023:simple}.
    In this decoupled framework, weight-free feature propagation is efficiently precomputed through the sparse matrix, eliminating time-based neighborhood sampling and gradient updates.
    This framework offers a potentially scalable solution for DGNNs but presents the following two significant challenges:
    C1: Due to the static topology (i.e., one timestamp) in simple graphs, traditional decoupled methods directly achieve graph propagation.
    However, the dynamic topology of DGs makes it impractical to pre-process node features at each individual timestamp due to expensive computational overhead.
    C2: Although traditional decoupled methods apply various propagation rules to different nodes~\cite{10.5555/gdc, huang2023node, li2024_atp}, they fail to dynamically capture the evolving temporal dependencies between nodes and their historical neighbors.

    To address C1, we propose Time-Aware Topology Reformulation (TTR), which reorganizes historical interactions using a two-step time partitioning principle to achieve fine-grained and weight-free dynamic graph propagation.
    Notably, previous studies segment the time range of historical interactions into discrete steps with equal intervals, similar to snapshot-based methods. 
    However, they oversimplify each step into a single timestamp, leading to inevitable information loss.
    Therefore, the key motivation of TTR is to incorporate finer-grained historical interactions within each timestamp. 
    Specifically, we reformulate temporal message passing from historical neighbors to the current time into intra-step propagation and inter-step propagation.
    In intra-step propagation, messages are propagated to the last timestamp of the step to form intermediate messages. 
    In inter-step propagation, the intermediate messages from all steps are transferred to the current time message. 
    This reformulation enables efficient intra- and inter-step propagation within a single time in preprocessing.

    To address C2, inspired by the widely used exponential functions in dynamic modeling, we introduce Dynamic Temporal Encoding (DTE) to further enhance TTR.~\cite{zuo2018:embedding,wen2022:trend,chanpuriya:2022direct}. 
    Notably, existing fixed exponential functions inadequately capture dynamic temporal dependencies. 
    Therefore, to better model complex temporal patterns, we propose DTE, which combines exponential functions with varying parameters to effectively model the influence decay of historical interactions.
    The key motivation of our approach is that, beyond the effectiveness of DTE in modeling temporal relationships, its multiplicative property enables seamless integration of intra- and inter-step interactions.
    Specifically, since nodes exhibit different dynamics across temporal states, determining the weights for these composite functions is critical. 
    We demonstrate that DTE, by applying a learnable transformation to the current message, is equivalent to performing an adaptive weight fusion of composite functions.
    However, simple transformations of the propagated features obtained by DTE-enhanced TTR result in all nodes sharing the same weights across different temporal states. 
    To address this issue, we introduce Hypernetwork-driven Message Aggregation. 
    The key motivation behind our method is to leverage hypernetwork~\cite{ha2016:hypernetworks,tay2020:hypergrid} to generate tailored transformation networks for each node, ensuring effective dynamic modeling of node-wise temporal dependencies.

    \textbf{Our contributions}.
    (1) \textit{\underline{New Perspective.}} 
    In this paper, we address the scalability challenges of existing DGNNs by proposing a novel time-oriented decoupled learning paradigm.
    (2) \textit{\underline{New Method.}}
    We propose ScaDyG, which first employs TTR to achieve weight-free dynamic graph propagation by two-step time partitioning. 
    This process can be efficiently computed using sparse matrix multiplication and is executed only once during pre-processing. 
    To further highlight temporal dependencies between nodes, we introduce DTE, which seamlessly integrates time influence in both inter and intra-step interactions. 
    Subsequently, through Hypernetwork-driven Message Aggregation, we achieve node-wise adaptive representation.
    (3) \textit{\underline{SOTA Performance.}}
    We conduct experiments on 12 datasets, including million-level dynamic graphs.  Results on link prediction and node affinity prediction tasks demonstrate that our approach achieves superior or comparable performance to SOTA baselines. Meanwhile, ScaDyG enjoys higher efficiency, with training times up to 60x faster and requiring up to 50x fewer parameters.

% Our contributions are as follows: 1) In this paper, we recognize the scalability challenges faced by existing dynamic GNNs and propose a novel paradigm for learning on large-scale dynamic graphs. 2)  We introduce a decoupled framework-based dynamic GNN called ScaDyG. We first employ a two-step dynamic feature propagation framework, leveraging a parameter-free sparse matrix for efficient feature propagation. Secondly, we propose a Dynamic Temporal Encoding  for temporal dependency modeling. It  seamlessly integrate the time influnce in inter and intra-step.  Finally, through Hypernetwork-driven
% Message Aggregation, we ensure node-wise modeling of temporal depencencies.   3) Extensive experiments on 13 real-life dynamic graph datasets demonstrate  superior or comparative performance to state-of-the-art baselines, with training times up 60x faster and requiring up to 50x less parameters.

\section{Related works}
\label{gen_inst}
\textbf{Dynamic Grpah Neural Networks.} The existing literature on dynamic graph neural networks can be categorized into discrete-based and continuous-based approaches.
% Dynamic graph neural networks are designed to  analyze the time-evolving graph data, such as constantly forming interactions in social networks 
 % \cite{panzarasa2009:patterns,bai:9378444,paranjape:2017motifs}, evolving connections between researchers in academic collaboration networks \cite{leskovec2005:graphs,} , and transaction networks \cite{}. 
 Discrete-based approach partitions a temporal graph into a series of snapshots with fixed
 time intervals. Typically, they are equipped with mechanisms dedicated to encoding structural patterns, such as GCN, within the snapshot as well as temporal dynamics, such as RNN, across snapshots \cite{hajiramezanali2019variational,yang2021:discrete,seo2018:structured,sankar2020:dysat,sun2021:hyperbolic,pareja2020:evolvegcn,you2022:roland,zhu2023:wingnn}.  Despite their simplicity, these methods overlook the temporal information within snapshots. SimpleDyG \cite{wu2024feasibility} is the only method that considers the order of interactions within a snapshot, focusing on fine-tuning a Transformer model for interaction sequences modeling.
%For example, some techniques utilize static graph neural networks to capture the structural information within individual snapshots, while leveraging sequence models such as RNNs \cite{hochreiter1997:long} or Transformers \cite{vaswani2017:attention} to encapsulate the dynamic characteristics across snapshots . 
Continuous-based methods model temporal graphs directly from the finer-grained temporal interactions.  They first sample a set of historical interactions from the neighborhood of a given node and aggregate the interactions with carefully designed encoders to compute the temporal embeddings. For instance, temporal random walk-based methods \cite{wang2021:inductive,jin:2022neural} model the propagation of temporal information through the process of random walks. Temporal encoding-based methods develop message massing-based methods temporal directly into message passing schemes \cite{xu2020:inductive,cong2023:we}. Moreover, temporal point process based methods \cite{zuo2018:embedding,trivedi2019:dyrep,lu2019:temporal,wen2022:trend} consider excitation effects of historical interactions to the occurrence of current interaction. However, the number of historical neighbors often increases rapidly with time, leading to a temporal explosion of interactions and scalability challenges. To enhance the efficiency of dynamic GNNs on large-scale DGs, several approaches \cite{li2023:zebra,zhou2022:tgl} propose efficient neighborhood sampling algorithms to accelerate the aggregation of neighborhood messages. Orca \cite{li2023orca} suggests reusing previously computed embeddings to reduce redundant computations. However, these methods still operate within the sampling-aggregating framework. 

%Different from existing these methods, we adopt a decoupled framework that fundamentally eschews the significant overhead of sampling and learnable neighborhood aggregation.
%Additionally, some methods view link prediction at different times as separate tasks and adopt a meta-learning approach by training a meta model to serve as the initial model, which is then fine-tuned on each snapshot \cite{}. 
% some approaches take into account self-excitation and mutual-excitation processes\cite{zuo2018:embedding,sankar2020:dysat,lu2019:temporal}, as well as distribution shifts on dynamic graphs\cite{zhang2024spectral,zhang2022:dynamic}.
%However, most of these approaches employ complex designs to characterize  temporal-spatial patterns,

\noindent
\textbf{Scalable Graph Neural Networks.}  Scalable Graph Neural Networks aim to process large-scale graph data by reducing the computational overhead during model training and inference.  Existing methods can be categorized into sampling-based approaches \cite{hamilton2017:inductive,chen2018fastgcn,chiang2019:cluster,zeng2019graphsaint,huang2018adaptive,zou2019:layer} and decouple-based approaches \cite{wu2019:simplifying,frasca:2020sign,zhu2020:simple,zhang2021:node,huang2023node}, while we primarily focus on the latter line of work. As the seminal decoupled GNN, Simplified Graph Convolution (SGC) \cite{wu2019:simplifying} first computes the node feature matrix results with feature propagation for multiple hops. The prediction results on downstream tasks, e.g., node classification are achieved by a logistic regression classifier based on node feature.  However, it employs a fixed receptive field for each node,  posing a limitation to adaptively leverage multi-hop neighborhood information. Therefore, subsequent works enhanced the node propagation rules by incorporating layer-wise propagation \cite{frasca:2020sign,zhu2020:simple} and node-wise propagation \cite{zhang2022:graph,zhang2021:node,huang2023node,liao2022:scara}. Other methods designed decoupled GNNs on sophisticated types of graphs, such as heterogeneous graphs \cite{yang2023:simple}, directed graphs \cite{li2024:lightdic} and user-item interaction graphs \cite{he2020:lightgcn}. Notably, TDLG \cite{chanpuriya:2022direct} is the only decoupled-based temporal graph embedding method in the literature. It models temporal associations between interactions using a line graph and exclusively focuses on edge-level tasks. However, its fixed temporal modeling function struggles to capture complex dynamics effectively. Overall, designing a scalable and effective framework for large dynamic graphs remains an open challenge.

% $k$ steps propagation  as $\mathbf{X}^{(k)} = \mathbf{S}^{k} X^{(0)}$, where $\mathbf{S} = \mathbf{D}^{-\frac{1}{2}}(\mathbf{A+I})\mathbf{D}^{-\frac{1}{2}}$ is the normalized adjacent matrix, and $\mathbf{X}^{(0)}$ is the initial node feature matrix. 

\section{Preliminaries}
\subsection{Dynamic graph representation learning}

A dynamic graph can be characterized by  \( \mathcal{G} = (\mathcal{V}, \mathcal{E}, \mathcal{T}, \mathcal{X}) \), where \( \mathcal{V}\) represents the set of nodes, \(\mathcal{E} \) denotes the set of edges, \( \mathcal{T}\) is set of the the timestamps of all edges, and $\mathcal{X} = \{\mathbf{X}_v,\mathbf{X}_e\}$ is the feature matrix, including $d_v$ dimension node features and $d_e$ dimension edge features. Alternatively, it can be conceptualized as a chronologically ordered series of interactions \( \mathcal{I} = \{(u_i, v_i, t_i)\}_{i=1}^k \), in which each triplet \( (u_i, v_i, t_i) \) signifies the establishment of edge \( (u_i, v_i) \) at  \( t_i \), with \( k \) representing the total count of interactions. Dynamic graph representation learning aims to derive a mapping function that, for any given timestamp \(t\), leverages the accumulated information up to \(t\) to project nodes into their respective time-aware embeddings. These embeddings can be utilized for downstream link-level and node-level tasks.

\noindent
\textbf{Temporal message passing.} Dynamic graph neural networks have become the SOTA methods for dynamic graph representation learning. Most existing dynamic GNNs adhere to the temporal message-passing mechanism \cite{li2023:zebra}. Despite their diversity, most 1-hop temporal message passing algorithm is formulated as follows: 
% \begin{equation}
% \begin{aligned}
%     \mathcal{N}_u^{(t)} & = \{(u, v, t') | t'<t\}, \\
%     \mathbf{m}_{v}^t & = {\rm Message}(\mathbf{x}_{v}, t-t'), \\
%     \mathbf{h}_u^t & = {\rm Aggregate}\{\mathbf{m}_{v}^t, v \in \mathcal{N}_u^{(t)}\},
% \end{aligned}
% \end{equation}
\begin{equation}
    \mathbf{x}_{m,v}^t = \text{Message}(\mathbf{x}_{v}, t - t'), \quad \mathbf{h}_u^t = \text{Aggregate}\{\mathbf{x}_{m,v}^t \mid v \in \mathcal{N}_u^{(t)}\},
\end{equation}
where $\mathcal{N}_u^{(t)}= \{(u, v, t') | t'<t\}$ is the set of historical neighbors at current time, $\mathbf{x}_{m,v}^t$ is the temporal message computed by the temporal edge $(u, v, t')$, $\mathbf{x}_{v}$ is the edge feature of  the temporal edge  $(u, v, t')$, and $\mathbf{h}_u^t$ is the time aware embedding of $u$ at $t$.

% \subsection{Decoupled  Graph Neural Networks}
% The decoupled based approach achieves scalable graph learning by separating the process of feature propagation and  learnable transformations. The feature propagation is achieved in the preprocessing stage through efficient sparse matrix computation, thereby  reducing the computational overhead of training the model.

\begin{figure*}[t!] 
  %\centering
   \vspace{-0.2cm}
  \includegraphics[width=\textwidth]{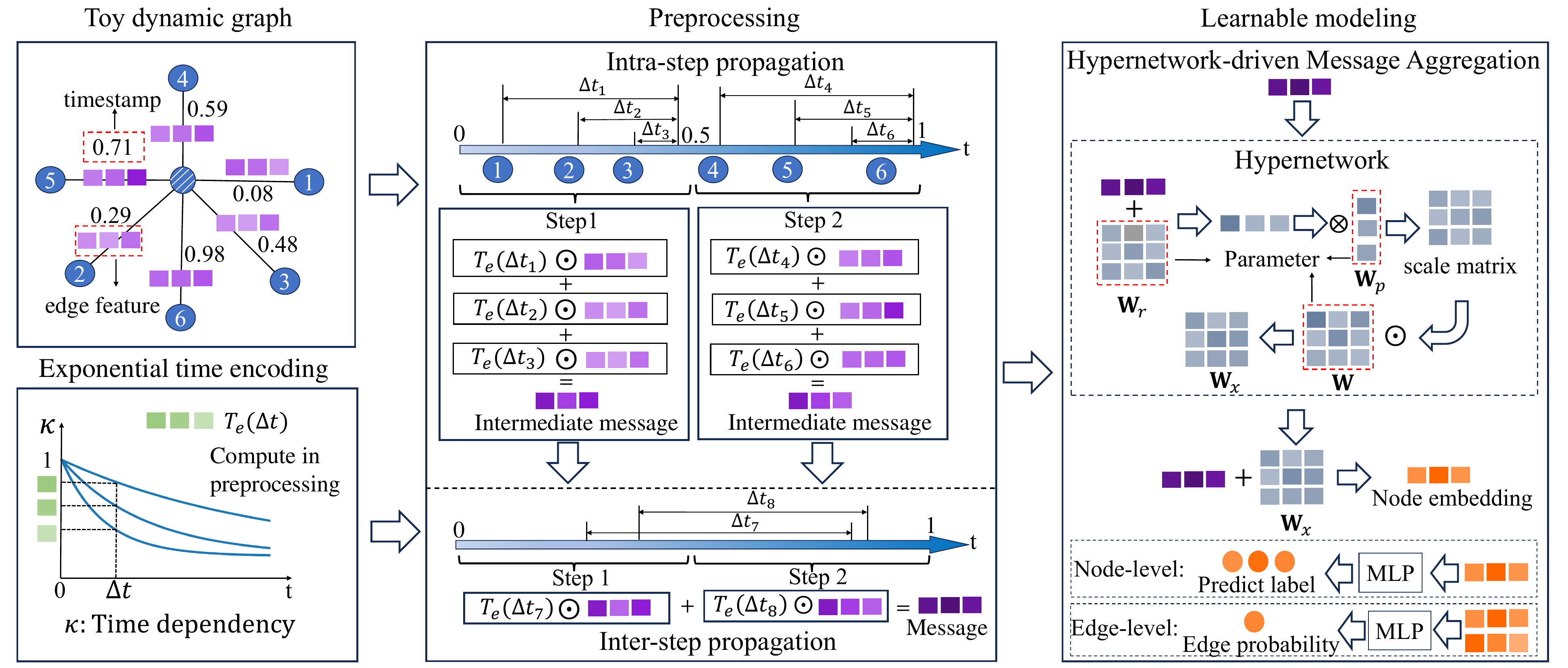}
 \vspace{-0.8cm}
  \caption{Overview of  ScaDyG. $\odot$ denotes the element-wise multiplication, and $\otimes$ denotes the outer product. We display the propagation from edge-to-node features in two steps as a toy example. }
  %\vspace{cm}
\label{fig:architecture}
\vspace{-0.4cm}
\end{figure*}
% Node features should  be first propagated to edge features, please refer to section \ref{TTR} for detailed implementations.

\section{The Proposed Method}
    
  In this section, we first introduce Time-aware Topology Reformulation, which computes temporal message passing in a decoupled framework. We decompose message computation into intra- and inter-step propagation, utilizing exponential time encoding as a unified time modeling in two phases.  In dynamic time encoding,  we demonstrate that  ScaDyG can dynamically model the time dependency with exponential time encoding. Finally, we present Hypernetwork-driven Message Aggregation. By utilizing a hypernetwork to scale the transformation matrix, we achieve node-wise message passing across different temporal states. The overview of ScaDyG is shown in figure \ref{fig:architecture}.

  \subsection{Time-aware Topology Reformulation}
  \label{TTR}
    %\textbf{Dynamic subgraph . } 
    \textbf{Decoupling of temporal messages.} \textcolor{black}{The core idea of Time-aware Topology Reformulation is to separate the message-passing process with steps, decoupling the fine-grained intra-step interactions modeling from the coarse-grained inter-step modeling. Specifically, the steps represent a series of equal time intervals covering the time span of historical interactions. For simplicity, all nodes are divided using the same interval. We denote the  steps as $s_1,s_2,...s_L$, which are divided by the series of timestamps $[t_{s_1}, t_{s_2}, \ldots, t_{s_L}]$.   The $i$-th  step is defined as  $(t_{s_{i-1}},t_{s_i}]$, where $t_{s_i}$ is the largest timestamp of the $i$-th step. Using the steps, the time range for computing a temporal message, i.e., $[t',t)$ can be split into two parts. Suppose $t'$ is in the $i$-th step, the two parts are $[t',t_{s_i})$ and $[t_{s_i},t)$, separately. Our idea is to first compute the message in $[t',t_{s_i})$ with preprocessing, obtaining an intermediate message, and then compute the message passing in $[t_{s_i},t)$ with the intermediate messages.   For example, in Figure 1, if we need to compute the central node embedding at both 0.5 and 1 timestamps, continuous-time methods would compute the messages of nodes 1, 2, and 3 twice, since they are the historic neighbors at both timestamps. Our method computes the message for nodes 1, 2, and 3 only once, storing it as an intermediate message at time 0.5. At time 1, we simply reference this intermediate result and combine it with the message of nodes 4, 5, and 6, reducing the unnecessary re-computation. We denote the first phase as intra-step propagation and the second as inter-step propagation.}
    %Finally, the propagated results are aggregated in a learnable manner.  
    %

% We denote the time steps split by equal  time intervals as \(\mathcal{I}_p = \{I_{t_1}, I_{t_2}, \ldots, I_{t_n}\}\), where each element $I_{t_s}$ is a sequence of temporal edges within the time range $(t_{s-1},t_{s}]$. $t_s$ is the largest edge timestamp in each sequence $I_{t_s}$. 
% \noindent
% \textbf{Decomposition of temporal message.} We denote the message passed from a historical neighbor \( v \) to \( u \) at current time \( t \) is represented as \( \mathbf{m}_v^t =  \mathbf{h}_v \theta (t-t')\mathbf{W} \) , where $\mathbf{W}$ is a learnable parameter matrix and $\theta$ is a time encoding function, $t'$ is the time stamp of the interaction, which fall into the $i$ the step, i.e. $t_{i-1} < t' \leq t_i$. 
% %The final node embedding at $t$ is computed through the aggregation of messages. 
% We formulate the computation of \( \mathbf{m}_v^t \) into intra-step propatation and inter-step transfer: the intra-step propatation spans from \( t' \) to $t_i $, and the inter-step transfer from \( t_i \) to \( t \). In the intra-step propatation, we obtain a intermediate feature at time \( t_i \), denoted as \( \mathbf{m}_v^{t_i} \). All intermediate features  are then  aggregated  in the inter-step transfer to obtain $\mathbf{m}_v^t$, and then compute the  node representation of $u$. Feature propagation from historical  neighbors only involves in the intra-step propagation, which is parameter-free. The inter-step transfer  exclusively focus on learnable temporal modeling.

\noindent
\textbf{Intra-step propagation.} Intra-step propagation computes message passing within each step, preserving the temporal information of each interaction, which is oversimplified in discrete-based methods.  Specifically, let $ t_j $ denote the timestamp of the $j$-th temporal edge in the $i$-th step. The time interval between \( t_{j} \) and the last timestamp  of the step  is $ \Delta t_{j} =    t_{s_i} -  t_{j} $.  Therefore, the time intervals of all temporal edges in the step are denoted as $ \Delta t_1, \ldots, \Delta t_{k_i}$, where $k_i$ is the number of edges in the step. To encode these time intervals, we introduce an exponential time encoding in Definition \ref{def1}.

\begin{definition}(Exponential time encoding).
Given a set of predefined parameters $\gamma_1,\gamma_2...,\gamma_{d_e} $ and a time interval $\Delta t$ ,  the exponential time encoding of $\Delta t$ is defined as $T_e(\Delta t) = [e^{\gamma_1\Delta t},e^{\gamma_2\Delta t},...,e^{\gamma_{d_e}\Delta t}]$.
\label{def1}
\end{definition}
We will discuss the effectiveness of this encoding in section \ref{sec:DTE}.   The temporal features encoded by $T_e$ are represented by $\mathbf{X}_t \in \mathbb{R}^{k \times d_e}$. To incorporate the edges with time information, these time features are element-wisely multiplied with edge features: $\mathbf{X}_e^{t_{s_i}} = \mathbf{X}_t \odot \mathbf{X}_e^i$. Next, we design two operators, $\mathbf{A}_{e,v}$ and $\mathbf{A}_{v,e}$, to propagate edge features to nodes and node features to edges, respectively.  Specifically, $\mathbf{A}_{e \rightarrow v} \in \mathbb{R}^{|\mathcal{V}|\times k_i}$ is the edge-node adjacent matrix where each element $A_{v_i,e_j}$ is defined as 
% On dynamic graphs, features are mainly associated with edges. Therefore, it is necessary to design operators that  propagate features from neighboring edges and  nodes to the given node. For a dynamic graph $\mathcal{G}_s$, two  operators $\mathbf{A}_{e,v}$ and $\mathbf{A}_{v,e}$ are designed to propagate edge features to nodes and node features to edges, respectively. Note that we've omitted the superscripts $(s)$ for simplicity, which denote that the letters are specific to the subgraph $\mathcal{G}_s$.  $\mathbf{A}_{e,v} \in \mathbb{R}^{|\mathcal{V}|\times m_k}$ is the edge-node adjacent matrix where each element $A_{v_i,e_j}$ is defined as 
\begin{equation}
A[v,e] = 
\left\{ 
\begin{array}{cl}
 1 & \text{if edge } e \text{ is connected to node }  v \\
 0 & \text{if edge } e \text{ is not connected to node }  v.
\end{array}
\right.
\end{equation}
% where $Tem(\cdot,\cdot)$ is the temporal encoding function that projects the timestamp into a rate in $(0,1]$. 
$\mathbf{A}_{v\rightarrow e} \in \mathbb{R}^{k_i\times |\mathcal{V}|}$ is the node-edge adjacent matrix, obtained by transposing $\mathbf{A}_{e,v}$.  We've omitted the superscripts $(i)$ for simplicity, which denotes that the letters are specific to the step $s_i$.  Therefore, The 1-hop propagation of edge features to their connected nodes is constructed as $\mathbf{X}_{m,e} = \mathbf{A}_{e\rightarrow v}\mathbf{X}_{e}^{t_{s_i}}$, where the subscript $m$ denotes the message.   Here, each node's intermediate message is the sum of the features of the adjacent edges. The reason for using sum instead of mean is to preserve the effect of repeated interactions. \textcolor{black}{To incorporate node features,  the initial node features are first propagated to connected edges and then to adjacent nodes, which is computed as $\mathbf{X}_{e,v} = \mathbf{A}_{v\rightarrow e} \mathbf{X}_v$. }Time is not involved in this process because we just want to transform edge features to node features. Next, we element-wisely multiply $\mathbf{X}_{e,v}  $ with the time feature and then propagate it to the nodes using the edge-node matrix $A_{e \rightarrow v}$ as previously described, forming $\mathbf{X}_{m,v}$.  Finally, we concatenate $X_{m,e}$ with $X_{m,v}$ to derive the intermediate messages $\mathbf{X}_{m} = [\mathbf{X}_{m,e}||\mathbf{X}_{m,v} ]$.

The design of these two operators is based on two insights.  Firstly, in dynamic graphs, temporal information is associated with edges, making edge-to-node propagation essential for integrating temporal information into nodes.  Secondly,  since a node could form edges with another node multiple times, node features should be propagated through each edge to preserve each individual interaction. Therefore,  for node-to-node propagation, node features should first propagate to their associated edges and then to the adjacent nodes. 

\noindent
\textbf{Multi-hop propagation.} Our framework can conveniently support multi-hop features. To this end, we follow  \cite{wu2019:simplifying} by first performing multi-hop propagation, and then concatenating the results of each hop's propagation. Specifically, the $l$-hop node and edge features are iteratively computed as   $\mathbf{X}^{(l)}_{m,v} =  \mathbf{A}_{e\rightarrow v}\mathbf{A}_{v\rightarrow e}\mathbf{X}_{v}^{(l-1)}$ and $\mathbf{X}^{(l)}_{e} = \mathbf{A}_{v\rightarrow e}\mathbf{A}_{e\rightarrow v}\mathbf{X}_{m,e}^{(l-1)}$.  We provide results and analysis of multi-hop propagation in section \ref{param}.

% After obatining  the node features for all historical  steps, i.e., from   the current node representation is computed through historical time-step aggregation.

\noindent
\textbf{Inter-step propagation.}  Given the intermediate messages at each step, the inter-step propagation first models the time interval from each step to the current time using $T_e$. Then, it aggregates the intermediate messages from all steps to obtain a current message. We denote intermediate messages obtained at all steps, i.e., $s_1,...,s_L$, as $\mathbf{X}^{1}_m,...,\mathbf{X}^{L}_m$, where $L$ is the largest step so that $t_L \leq t$. The time interval between the timestamp of step $s_i$ and current timestamp is $\Delta t_i = t- t_{s_i}$, and then the time feature encoded is $T_e(\Delta t_i)$. To integrate the inter-step time intervals, the intermediate messages at each historical step $i$ are element-wisely multiplied by   $T_e(\Delta t_i)$, transferring them to the current time. To obtain the current messages, we  sum the transferred message across all steps: 
\begin{equation}
    \mathbf{X}_m^t = \sum_{i=1}^L \mathbf{X}^i_m.
    \label{transformation}
\end{equation}
\vspace{-0.5cm}
\subsection{Dynamic time encoding}
\label{sec:DTE}
In Time-aware Topology Reformulation, we introduced exponential time encoding as a general time modeling in intra and inter-step propagation. Here, we further demonstrate the time encoding can be as expressive as a dynamic fusion of diverse parameterized exponential functions with a simple learnable transformation. We begin by formulating the concept of composite exponential dependency. 

\begin{definition} (Composite exponential dependency).
    Given an exponential time encoding as $T_e(\Delta t) = [e^{\gamma_1\Delta t},e^{\gamma_2\Delta t},...,e^{\gamma_{d_e}\Delta t}]$ and a set of learnable parameters $[a_1,a_2,...,a_{d_e}]$. The composite exponential  dependency is defined as $\kappa (\Delta t) = a_1e_1^{\gamma_1\Delta t}+a_2e_2^{\gamma_2\Delta t}+...+a_{d_e}e^{\gamma_{d_e}\Delta t}$.
\end{definition}
Exponential functions have shown their effectiveness in modeling time dependency decay \cite{hawkes1971:spectra,wen2022:trend,zuo2018:embedding}. However, existing methods typically employ a fixed exponential function to model such effects of historical neighbors \cite{zuo2018:embedding,wen2022:trend,chanpuriya:2022direct}. A fixed $\gamma$ value fails to characterize the varying temporal patterns exhibited by different nodes and time states. In exponential time encoding, each element of the feature represents the time interval is modeled by a unique exponential function, and the dynamic fusion of the elements captures a diverse range of temporal patterns. Therefore, by applying a weighted sum of these exponential functions, composite exponential (CE) dependency achieves good representational capacities for modeling temporal dependencies \cite{zhou2013:learning}. Building on CE dependency, we introduce composite exponential (CE) message passing, a robust implementation of learnable message passing that utilizes CE dependency as the temporal modeling function.

\begin{definition}(Composite exponential message passing).
    Given an composite exponential dependency $\kappa (t-t')$ and a learnable matrix $\mathbf{W}$. Assuming a historical neighbor $v$ interacts with $u$ at $t'$, $t$ is the current timestamp, and the time interval $\Delta t = t- t'$.   Its composite exponential (CE) message passing is defined as $ {\rm CE} (\mathbf{x}_v,t-t') =  \mathbf{x}_v \kappa (t-t')\mathbf{W}$.
\end{definition}
Next, we demonstrate that ${\rm CE} (\mathbf{x}_v,t-t')$ can be computed sequentially in preprocessing and learnable modeling.   

\begin{proposition}
If $\Delta t = t-t'$ can be split by $t_s$ as $\Delta t_1 = t_s-t'$,  $\Delta t_2$ = $t - t_s$, and   $\mathbf{W}_1$ is a learnable parameter matrix,   $\mathbf{x}_v\odot T_e(\Delta t_1)\odot T_e(\Delta t_2)\mathbf{W}_1 $ is equivalent to $ \mathbf{x}_v \kappa (\Delta t) \mathbf{W}$.
 \label{prop1}
\end{proposition}

The proof of  Proposition \ref{prop1} is given in  Appendix \ref{proof}. From Proposition 1, we can observe  $\mathbf{x}_v\odot T_e(\Delta t_1)\odot T_e(\Delta t_2)$  involves no learnable parameter. When setting $t_s = t_{s_i}$, i.e., the largest timestamp of the $i$-th step,  $\mathbf{x}_v^{t_i} = \mathbf{x}_v \odot T_e(\Delta t_1)$ corresponds to the intra-step propagation. Conversely, $\mathbf{x}_v^t = \mathbf{x}_v^{t_{s_i}} \odot T_e(\Delta t_2)$ corresponds to inter-step propagation. In Time-aware Topology Reformulation, we sum all messages across nodes to obtain $\mathbf{X}_m^t$. By multiplying $\mathbf{X}_m^t$ with a learnable matrix $\mathbf{W}$, we effectively implement CE message passing from all historical neighbors. Although no learnable parameters is involved during preprocessing, our analysis shows that ScaDyG achieves dynamic modeling of temporal dependencies. 

\subsection{Hypernetwork-driven Message Aggregation}

Although we have established a decoupled framework dynamically modeling temporal dependencies, simple message aggregation with $\mathbf{W}$ has the following limitations.
1) It applies the same transformation for all nodes,  ignoring the dynamic differences between individual nodes. 2)   In the inference stage, $\mathbf{W}$ is frozen.  
This hinders the model to adapt and update with the evolving of new data. To address the limitations, we introduce a Hypernetwork-driven Message Aggregation, enabling node-specific and dynamic updates to the transformation matrix in equation \ref{transformation} as time evolves. 

A hypernetwork \cite{ha2016:hypernetworks} is a meta-model designed to generate weights for a main network. In our context, the main network refers to the transformation matrix for aggregating messages.  The idea for generating node-specific weights is to select node-related identifiers as inputs to the hypernetwork. The messages of nodes, i.e., $\mathbf{X}_s^t$ are natural choices for these identifiers, as it captures the historical neighborhood interactions that characterize the node's current state. To generate the weights, we apply gird-wise projections \cite{tay2020:hypergrid} for two reasons. 1) Fewer extra parameters. Only an additional matrix and a vector are introduced. 2) Good Interpretability.  
 Different node patterns and states are reflected in specific regions of projected weights, as we will show in section \ref{Interpretability}.

% aking inspiration from HyperGrid \cite{tay2020:hypergrid}, we introduce a weight scaling method for weight generation, which realizes node specific $\mathbf{W}$ at the cost of acceptable extra learnable parameters.

Specifically, we first introduce a primary weight matrix $\mathbf{W}$, and use a projection weight generated by hypernetwork to scale it to a node-specific matrix. To this end, we introduce two additional learnable parameters $\mathbf{W}_r \in \mathbb R^{d \times d}$ and $\mathbf{W}_p \in \mathbb R^{d}$ apart from $\mathbf{W}$. The aim of $\mathbf{W}_r$ is to generate row vector specific node features, while $\mathbf{W}_p$ is a column vector extending it into a scale matrix with the outer product. Finally, we perform element-wise multiplication with the scale matrix and $\mathbf{W}$, projecting it into a node-specific transformation matrix $\mathbf{W}_x$.  The projection  is formulated as:
\begin{equation}
    \mathbf{W}_x  = (\sigma(\small(\mathbf{W}_r\mathbf{X}^t \small) \otimes \mathbf{W}_p ))\odot \mathbf{W},
\end{equation}
where $\sigma$ is the sigmoid function, and $\otimes$,$\odot$ are the element-wise product and the outer product, respectively.   To obtain the final node representation, we aggregate the messages in equation \ref{transformation} with a feature transformation using $\mathbf{W}_x$  :
\begin{equation}
    \mathbf{X}^t  = \mathbf{X}_s^t\mathbf{W}_x.
\end{equation}

\begin{table*}[ht]
\centering
\small
\caption{The statistics of experimental dynamic graph datasets.}
\vspace{-0.2cm}
\label{your_label}
\resizebox{\textwidth}{!}{
\begin{tabular}{@{}cccccccccc@{}}
\toprule
Datasets & \# Nodes  & \# Edges  & \# Node/Edge features & \# Unique Timestamps & \# Time Steps & Description &  Task & \#Avg.Deg & \\ \midrule
UCI & 1,899 & 59,835 & -/- & 58,911  & 273 & Social & Link prediction &  31.51\\
% Wikipedia & 9,227 & 157,474 & -/172 & 152,757  & 100 & Interaction & Link prediction & 17.06\\
MOOC & 7,144 & 411,749 & –/4 & 345,600  & 100 & Interaction & Link prediction & 57.63\\
BitcoinAlpha & 3,783 & 24,186 & -/2 & 1,647 & 226 & Transaction & Link prediction  & 6.93\\
LastFM & 1,980 & 1,293,103 & -/- & 1,283,614  & 500 & Interaction & Link prediction  &653.08\\
UNvote & 201 & 1,035,742 & -/1 & 72  & 72 & Politics & Link prediction  & 5152.94\\
% Flights &  13,169 &  1,927,145 & -/1  & 122  & 122 & Transportation \\
Reddit-title & 54,075 & 571,927 & 300/88 & 354,507  & 178 & Interaction  & Link prediction & 10.57\\
Enron &  184 & 125,235 & -/- & 22,632 & 100& Social & Link prediction  & 680.63 \\ 
Stackoverflow & 2,601,977 & 63,497,050 & -/- &  1,846,553   & 91 & Interaction & Link prediction &24.20 \\
\midrule
tgbl-trade & 255 & 468,245 & -/1 & 30  & 30 & Economics & Node affinity prediction  & 1836.25\\
tgbl-genre & 1,505 & 17,858,395 & -/1 &  4,187,046& 1580 & Interaction & Node affinity prediction & 11,866.04\\
tgbl-reddit & 11,766 & 27,174,118 & -/1 &  21,889,537   & 1090 & Social & Node affinity prediction &2309.55\\
tgbl-token &  61,756 & 72,936,998 & -/1 & 2,036,524 & 785 & Transaction & Node affinity prediction & 1181.05\\
\bottomrule
\end{tabular}
}
\label{table_dataset}
\vspace{-0.2cm}
\end{table*}

% Through dynamic feature propagation, we can obtain node representation of the dynamic graph within the time range of each subgraph. Our goal is to aggregate historical states to obtain the representation of nodes at the current moment.  

\noindent
\textbf{Training objective.}  For node-level tasks, the node representations are fed into an MLP for generating prediction labels. We use the cross entropy based objective function as follows:

\begin{align}
   & \mathcal{L}_{node} = \rm \sum_{\mathit{i}} \sum_{u\in \mathcal{V}} -\mathbf{Y}_\mathit{vi}ln \left(Softmax \left(MLP(\mathbf{X}^{\mathit{t}})\right)_\mathit{ui}\right),
\end{align}
where the subscript $vi$ denotes the $i$-th label of node $u$. For link prediction, we replace $\mathbf{X}^t$ with the concatenation of source and target node representation, and sample a negative edge for each existing edge following \cite{yu2023towards}
to construct training pairs.
%\subsection{Scalable dynamic graph leaning} 
% In this section, we present the aforementioned dynamic feature propagation and temporal aggregation operations through a scalable framework. 
%Given a dynamic graph \(\mathcal{G}\), 
% we denote the  the set of temporal interactions split by  \(\mathcal{I}\) into \(n\) consecutive sequences, denoted as \(\mathcal{I}_p = \{I_{t_1}, I_{t_2}, \ldots, I_{t_n}\}\). 

% In this context, each sequence \(I_{t_s}\) consists of  a sequence of $m_s$ temporal edges \(\{(u,v,t) \mid t_{s-1} < t \leq t_{s}\}\). 
% The subgraph bulit upon  $I_{t_s}$ is denoted as $\mathcal{G}_s = \{\mathcal{V}, I_s,\mathbf{X}_v,\mathbf{X}_e^s\}$, where the node set and node feature matrix are shared across all subgraphs.
% $\mathbf{X}_e^s \in \mathbb{R}^{k_s \times d_e}$ is the edge feature matrix composed of the features corresponding to each temporal edge. Below, we formulate the two phases, i.e., the intra-time-step propagation and historical time-step aggregation of dynamic feature propagation with efficient matrix operations.  
%Please note that the inductive

% Unlike the snapshot-based method that treats each snapshot as a static graph, we preserve the temporal information of each edge within every subgraph. 

\

\vspace{-0.6cm}

\section{Experiments} 

    In this section, we conduct extensive experiments on our approach.
    To begin with, we introduce 13 benchmark datasets along with prevalent DGNN baselines. 
    Subsequently, we present the temporal-based evaluation methodology. 
    Details about these experimental setups can be found in Appendix \ref{detail}.
    After that, we aim to address the following questions:
    \textbf{Q1}: Compared to existing DGNNs, can ScaDyG achieve SOTA predictive performance?
    \textbf{Q2}: What is the running efficiency of ScaDyG, especially in large-scale scenarios?
    \textbf{Q3}: If ScaDyG is effective, what contributes to its performance?
    \textbf{Q4}: How robust is ScaDyG when dealing with hyperparameters?

\subsection{Experimental settings}

\noindent
\textbf{Datasets and Baselines.}  
    We conduct link- and node-level evaluations on 12 datasets from 5 real-world application domains.
    The statistics of these datasets are presented in Table~\ref{table_dataset}.
    For baselines, we compare ScaDyG with the following baselines:
    (1) Discrete-based methods: EvolveGCN \cite{pareja2020:evolvegcn} and ROLAND \cite{you2022:roland});
    (2) Continuous-based methods: DyGFormer \cite{yu2023towards}, GraphMixer \cite{cong2023:we}, TGAT \cite{xu2020:inductive}, TGN \cite{rossi2020:temporal}, DyRep \cite{trivedi2019:dyrep}, \textcolor{black}{JODIE} \cite{kumar:predicting}.
    Detailed descriptions of the above datasets and baselines are provided in Appendix~\ref{datasets}-\ref{baselines}.

%the model forecasts whether a link will form between two nodes at a future timestamp. 
\noindent
\textbf{Evaluation.} 
    \textcolor{black} {Regarding link-level prediction, we select transductive link prediction, a widely used task in existing studies \cite{yu2023towards,cong2023:we,xu2020:inductive,rossi2020:temporal,you2022:roland}.
    In link prediction,  we follow \cite{you2022:roland} to use the ranking setting and MRR as the evaluation metric. We also report  AP and AUC in binary classification setting in Appendix ~\ref{add_results}.
    As for node-level experiments, we choose the recently introduced node affinity prediction \cite{huang2024:temporal},   where the model predicts a multi-dimensional label representing the affinity between a node and a set of nodes. We use NDCG as the evaluation metric.} We provide detailed background and motivation for the tasks, metrics and datasets in Appendix \ref{detail}.

\begin{table*}
    {\small% Begin small font size environment
    \caption{Performance on dynamic link prediction. The best result is bold, the second best result is underlined. \rm OOM denotes that the method encountered an out-of-CUDA-memory issue, while OOT indicates that the training did not finish within 48 hours.
}
\vspace{-0.2cm}
\resizebox{\textwidth}{!}
{\begin{tabular}{c|c c c c c c c c c c }
\hline
Datasets & JODIE & EvolveGCN & ROLAND & DyRep & TGAT & TGN & GraphMixer & DyGFormer & ScaDyG  \\ % First row
\hline
UCI &0.402 $\pm$ 0.003  &0.213 $\pm$ 0.009 & 0.419 $\pm$ 0.014 &0.434 $\pm 0.007$
 &0.515 $\pm$ 0.016
 & 0.463 $\pm$ 0.011
&0.522 $\pm$ 0.008
 & \textbf{0.601} $\pm 0.031$ & \underline{0.534} $\pm$ 0.007\\ % Second row

% Wikiepdia &0.367 $\pm$ 0.005 &0.356 $\pm$  0.008 &0.756 $\pm$ 0.017 &  $\pm 0.020$
% &0.780 $\pm$ 0.007
%  & \underline{0.855} $\pm$ 0.010
%  & 0.799 $\pm$ 0.016
% &$\pm$ 0.012
%  & \textbf{0.931} $\pm$ 0.009 \\ % Third row
MOOC & 0.356 $\pm$ 0.014 &0.373 $\pm$ 0.016 &0.756 $\pm$ 0.017 & 0.653 $\pm 0.020$
&0.780 $\pm$ 0.007
 & \underline{0.855} $\pm$ 0.010
 & 0.799 $\pm$ 0.016
&0.730 $\pm$ 0.012
 & \textbf{0.931} $\pm$ 0.009 \\ % Third row
BitcoinAlpha &0.332 $\pm$ 0.011 &0.157 $\pm$ 0.022 &0.232 $\pm$ 0.023 & 0.174 $\pm$ 0.005
 &0.319 $\pm$ 0.023  & 0.218 $\pm$ 0.006
 & 0.224 $\pm$ 0.026
&\underline{0.503} $\pm$ 0.030 & \textbf{0.597} $\pm$ 0.019
 \\ % Fourth row
LastFM & 0.098 $\pm$ 0.007 & 0.039 $\pm$ 0.006 & 0.053 $\pm$ 0.002 &0.159 $\pm 0.006$ & 0.193 $\pm$ 0.012

 & 0.042 $\pm$ 0.001& 0.177 $\pm$ 0.009

& \underline{0.236} $\pm$ 0.025 & \textbf{0.248} $\pm$ 0.012 \\ % Fifth row
UNvote &0.056 $\pm$ 0.019 & 0.405 $\pm$ 0.018 & 0.474 $\pm$ 0.010 & 0.446 $\pm$ 0.031 & 0.564 $\pm$ 0.018 & 0.596 $\pm$ 0.027 &0.519 $\pm$ 0.016 & \underline{0.657} $\pm$ 0.018
 & \textbf{0.725} $\pm$ 0.009 \\ % Sixth row
% Flights &0.397 $\pm$ 0.008 & 0.441 $\pm$ 0.016 & 0.429 $\pm$ 0.006
%  & \underline{0.706} $\pm$ 0.007&0.686 $\pm$ 0.013
%  &0.411 $\pm$ 0.013
%  & \textbf{0.778} $\pm$ 0.008
%  & 0.486 $\pm$ 0.009 \\ % Seventh row
% DBLP & & &0.108 $\pm$ 0.005 &0.066 $\pm$ 0.003 & 0.108 $\pm$ 0.003&0.095 $\pm$ 0.005 &0.530 $\pm$ 0.017 &  \\ % Eighth row
Reddit-title & 0.414 $\pm$ 0.019  &0.354 $\pm$ 0.020 & 0.440 $\pm$ 0.036  &0.537 $\pm$ 0.014
 &0.597 $\pm$ 0.011
 &0.628 $\pm$ 0.006
 &0.658 $\pm$ 0.015
 & \textbf{0.706} $\pm$ 0.010 & \underline{0.679} $\pm$ 0.013  \\ % Ninth row
Enron & 0.407 $\pm$ 0.006 & 0.061 $\pm$ 0.009 & 0.065 $\pm$ 0.005 &0.118 $\pm$ 0.007 &0.147 $\pm$ 0.010
 &0.235 $\pm$ 0.002
 &0.224 $\pm 0.008$ & \underline{0.433} $\pm 0.017$ &  \textbf{0.481} $\pm$ 0.009\\ % Tenth row
Stackoverflow &0.181 $\pm$ 0.016 &OOM & OOM &OOT  &OOT  &OOT  & \underline{0.231} $\pm$ 0.009 &0.217 $\pm$ 0.018 & \textbf{0.559} $\pm$ 0.012\\ % Tenth row
\hline
\end{tabular}
}\label{link_pred}
} % End small font size environment

\vspace{-0.2cm}
\end{table*}

\begin{table*}
    \caption{Performance on dynamic node affinity prediction. The best result is bold, the second best result is underlined.}
    \vspace{-0.2cm}
    \resizebox{\textwidth}{!}{
    {\small% Begin small font size environment
    \begin{tabular}{c|c c c c c c c c c c c c}
    \hline
    Datasets  &JODIE & EvolveGCN & ROLAND & DyRep & TGAT & TGN & GraphMixer & DyGFormer &   ScaDyG \\ % First row
    \hline
    tgbl-trade  & 0.373 $\pm$ 0.002  & 0.315 $\pm$ 0.011 & 0.439 $\pm$ 0.018 &0.376 $\pm$ 0.005 &\underline{0.448} $\pm$ 0.07 & 0.381 $\pm$ 0.006
 &0.366 $\pm$ 0.012
 &0.386 $\pm$ 0.008
 &\textbf{0.631} $\pm$ 0.007  \\ % Second row
    tgbl-genre & 0.331 $\pm$ 0.007 & 0.305 $\pm$ 0.019 &0.357 $\pm$ 0.013 &0.337 $\pm$ 0.011 & 0.243$\pm$ 0.009 & \underline{0.369} $\pm$ 0.021 & 0.330 $\pm$ 0.015 & 0.259 $\pm$ 0.018 & \textbf{0.403} $\pm$ 0.005 \\ % Third row
    tgbl-reddit & 0.327 $\pm$ 0.012 & 0.257 $\pm$ 0.017 & 0.334 $\pm$ 0.016 & 0.309 $\pm$ 0.011 & \underline{0.356} $\pm$ 0.014 &0.329 $\pm$ 0.018 &0.218 $\pm$ 0.015 &0.305 $\pm$ 0.027  & \textbf{0.402} $\pm$ 0.012 \\ % Fourth row
    tgbl-token &0.308 $\pm$ 0.010 & 0.234 $\pm$ 0.026 & 0.303 $\pm$ 0.018 & 0.159 $\pm$ 0.024&  \underline{0.334} $\pm$ 0.021 &0.185 $\pm$ 0.023 & 0.164 $\pm$ 0.022 & 0.251 $\pm$ 0.014  &\textbf{0.686} $\pm$ 0.019 \\ % Fifth row
    \hline
    \end{tabular}
    }
    }
    \label{node_property}
\vspace{-0.2cm}
\end{table*}

\subsection{Overall performance}

\noindent
\textbf{Link-level performance.} 
    To answer \textbf{Q1}, we first report the performance of ScaDyG in predicting temporal relationships between nodes on MRR, as shown in Table~\ref{link_pred}.
    Our findings indicate that ScaDyG consistently achieves the highest or second-highest performance on nine datasets, with an average improvement of 30.65\%, validating its effectiveness.
    Furthermore, we observe that discrete-based methods underperform compared to continuous-based methods because they neglect fine-grained temporal interactions within snapshots. 
    Meanwhile, discrete-based methods encounter OOM error on the StackOverflow dataset, while several continuous-based methods face OOT errors, indicating scalability struggles with complex models like RNNs or memory networks.
    Notably, Graphmixer and DyGformer perform poorly on the StackOverflow dataset, likely due to the large number of nodes introducing significant variability, which impairs their generalization ability and robustness.

\noindent
\textbf{Node-level performance.} 
    Subsequently, we use node affinity prediction to thoroughly evaluate the DGNNs' ability to capture dynamic relationships within node descriptions.
    Based on this, we report the overall performance in Table~\ref{node_property}.
    According to the experimental results, ScaDyG achieves the best performance across all datasets, with an average improvement of 42.05\%, highlighting its advantage in node-level inference. 
    Unlike several prevalent baselines that prioritize link-level optimizations, ScaDyG avoids task-specific optimizations and instead focuses on modeling dynamic dependencies of temporal interactions.
    This makes ScaDyG a more general framework for various DG-based downstream tasks.

\begin{table*}[]
\centering
\caption{Time and space complexity analysis of ScaDyG and existing dynamic graph learning models. \rm $m$:edge numbers, $n$: node numbers, $f$: feature dim (including node \& edge feature), $b$: the batch size, $N$: number of sampled neighbors for continuous-time based methods, and the number of historical steps in ScaDyG. $L$: number of total steps in ScaDyG and snapshots in EvolveGCN, $l$: layers of neural networks, and number of hops in ScaDyG. Link and node represent the complexity of the link and node-level tasks respectively.}
\vspace{-0.3cm}
   \begin{tabular}{c| c| c|c|c|c} 
 \midrule
Type& Model  &Pre-processing & Training/Inference (link) &Training/Inference (node) & Space \\ 
 \midrule
\multirow{2}{*}{\centering Discrete-time based} & EvolveGCN &  - & $O(Ll(n^2f+nf^2))$ & $O(Ll(n^2f+nf^2))$  & $O(nf+lf^2)$ \\  
 & ROLAND   & - & $O(l( m  +  n )f^2)$ & $O(l( m  +  n )f^2)$ & $O(nlf+lf^2)$ \\  
 \midrule
\multirow{5}{*}{\centering Continuous-time based} 
& JODIE & -  & $O(mlf^2)$ & $O(mlf^2)$ & $O((n+m)f+lf^2)$\\
& DyRep  & - & $O((n+mN)f^2)$ & $O(nf^2+mf^2+mN)$ & $O(nf+ bNf+f^2)$ \\  
 & TGN  & - & $O(mklNf^2)$  & $O(nklNf^2)$ & $O(nf+bNf+lf^2)$ \\  
 & TGAT  & - & $O(mkN^{l}f^{2})$ & $O(nkN^{l}f^{2})$  & $O(bN^lf+lf^2)$  \\  
 & GraphMixer  & - & $O(mNlf^{2})$ & $O(nNlf^{2})$ & $O(bNf+f^2)$ \\  
 & DyGFormer  & - & $O(mkNlf^2)$ & $O(nkNlf^2)$ & $O(bNf+lf^2)$ \\  
 \midrule
Decoupled based& ScaDyG  & $O(lmf)$ & $O(nNLf^2)$ & $O(nNLf^2)$  & $O(bLf+bf^2)$ \\ 
\midrule
\end{tabular}
\label{table_complexity}
\vspace{-0.3cm}
\end{table*}

\begin{figure*}
    \centering
    \begin{subfigure}{0.24\textwidth}
        \centering
        \includegraphics[width=\textwidth]{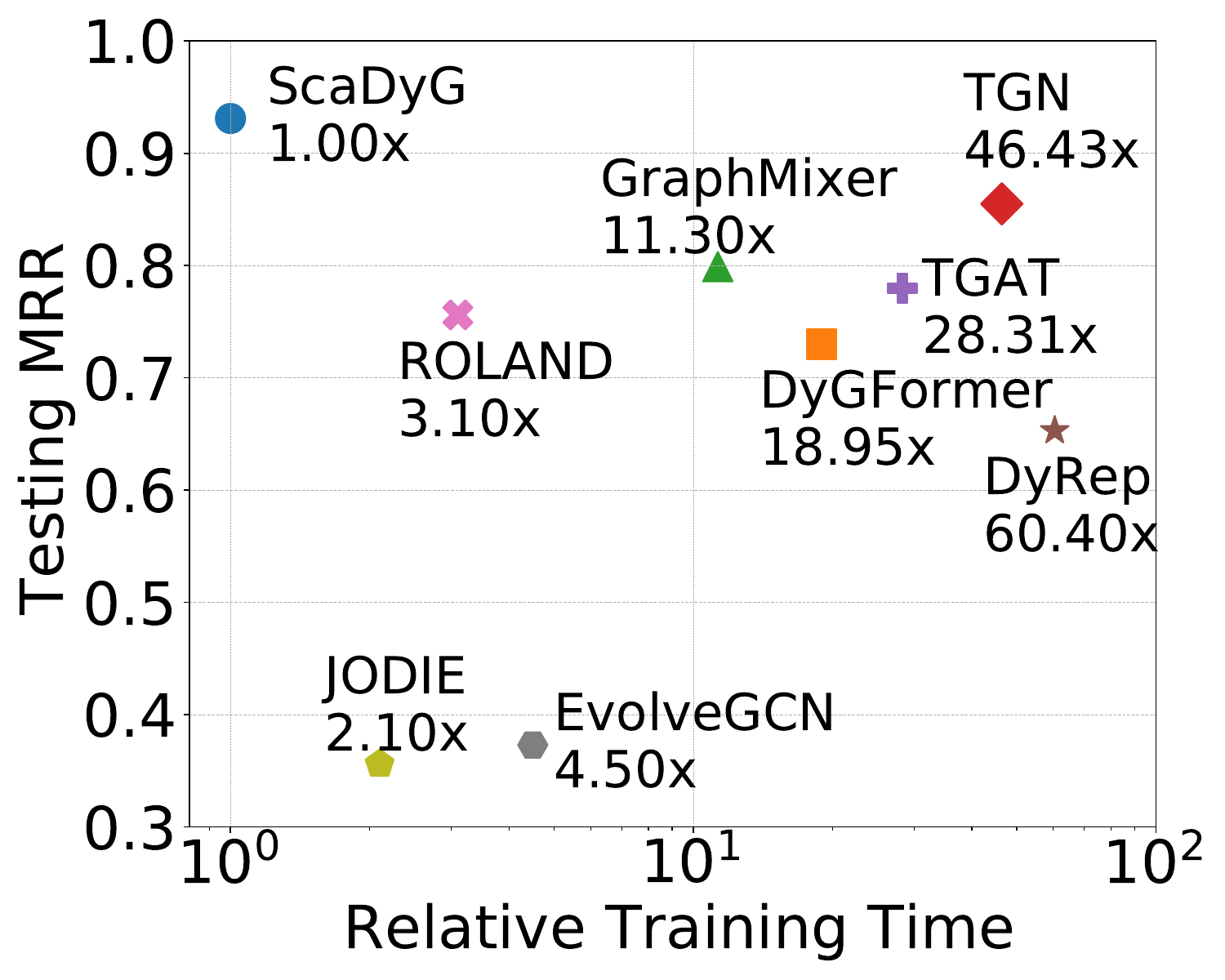}
        \caption{ Training time on MOOC}
        \label{fig:subfig1}
    \end{subfigure}
    \hfill
    \begin{subfigure}{0.24\textwidth}
        \centering
        \includegraphics[width=\textwidth]{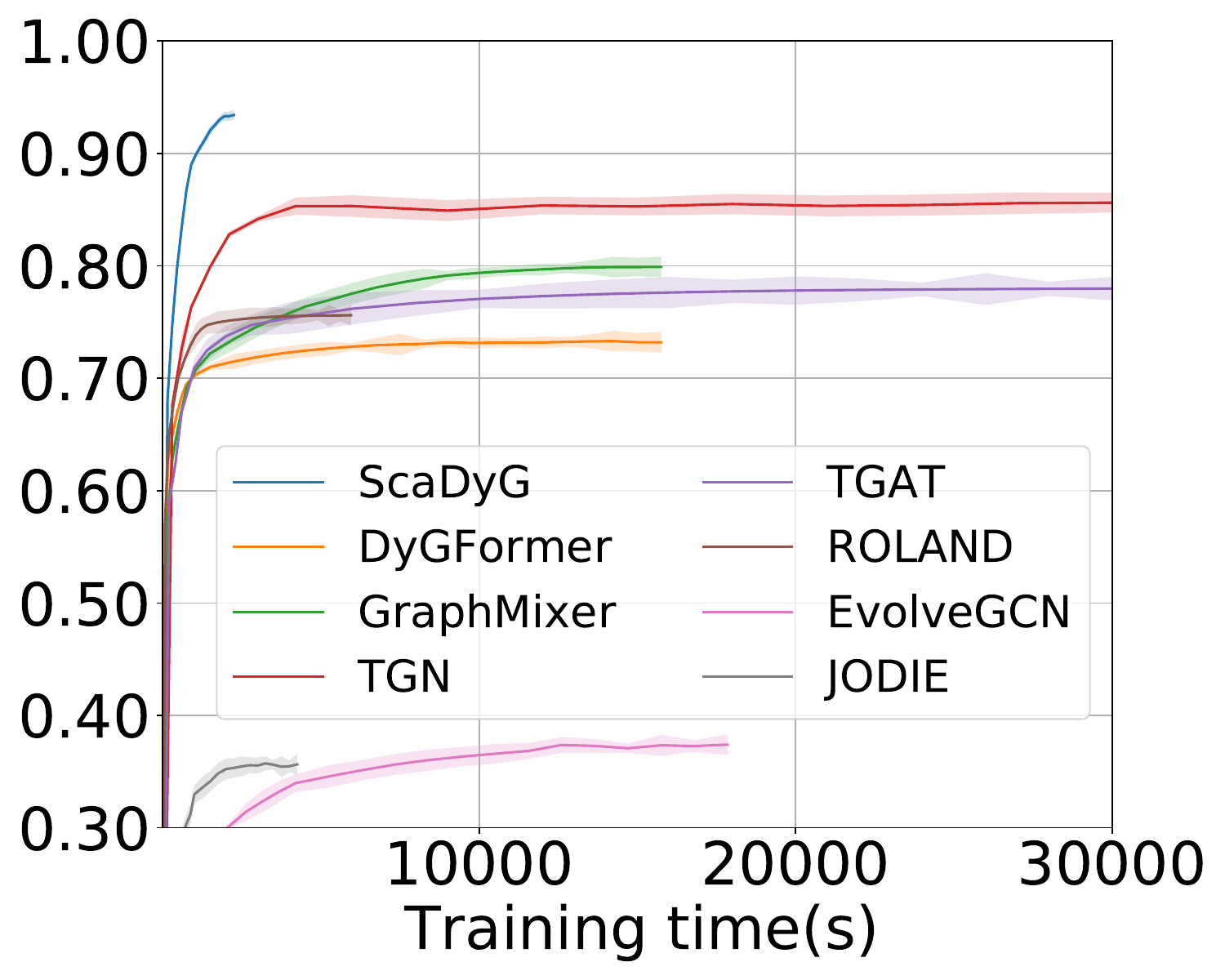}
        \caption{ Training time on MOOC}
        \label{fig:subfig2}
    \end{subfigure}
    \hfill
    \begin{subfigure}{0.24\textwidth}
        \centering
        \includegraphics[width=\textwidth]{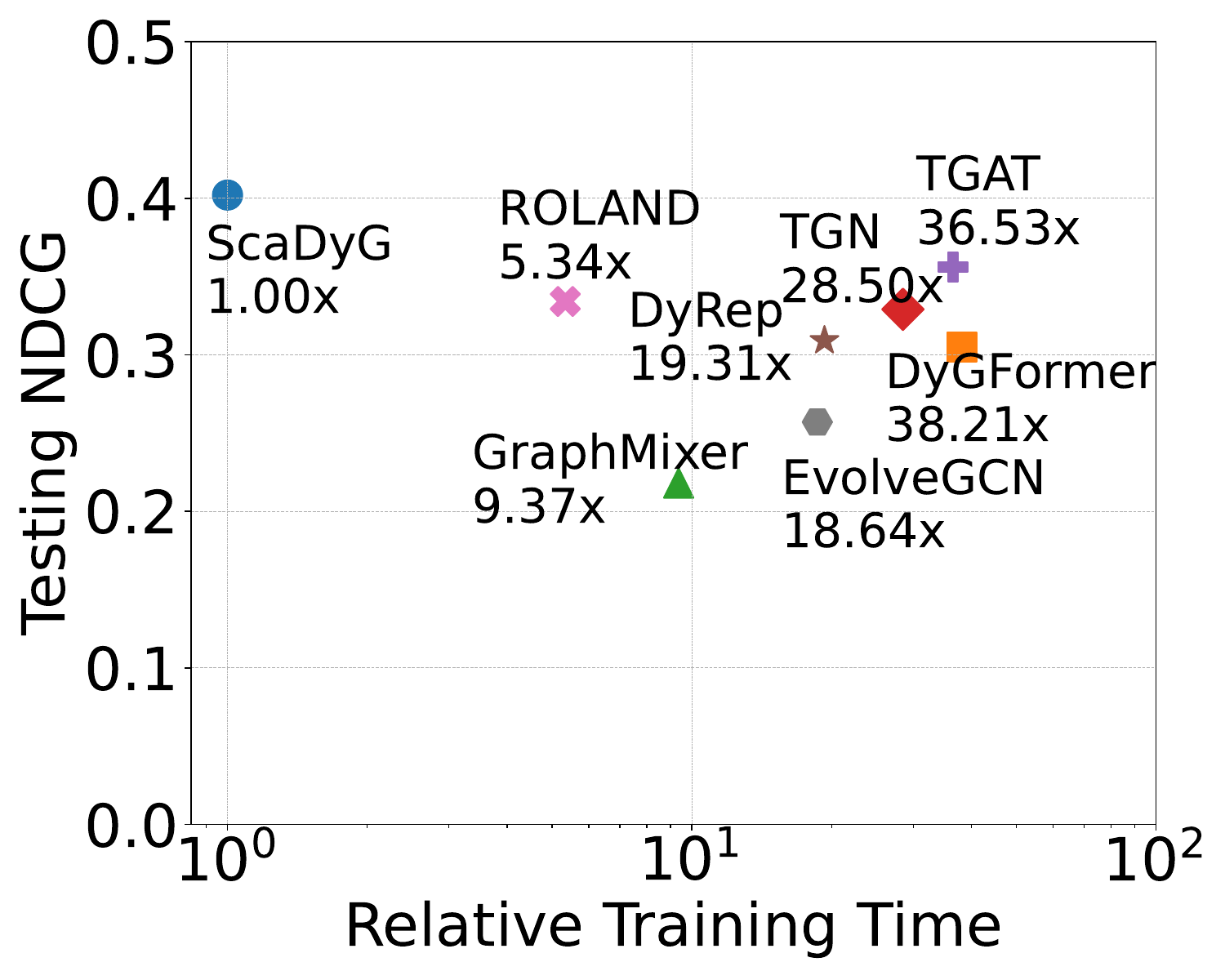}
        \caption{  Training time on tgbl-reddit}
        \label{fig:subfig3}
    \end{subfigure}
    \hfill
    \begin{subfigure}{0.24\textwidth}
        \centering
        \includegraphics[width=\textwidth]{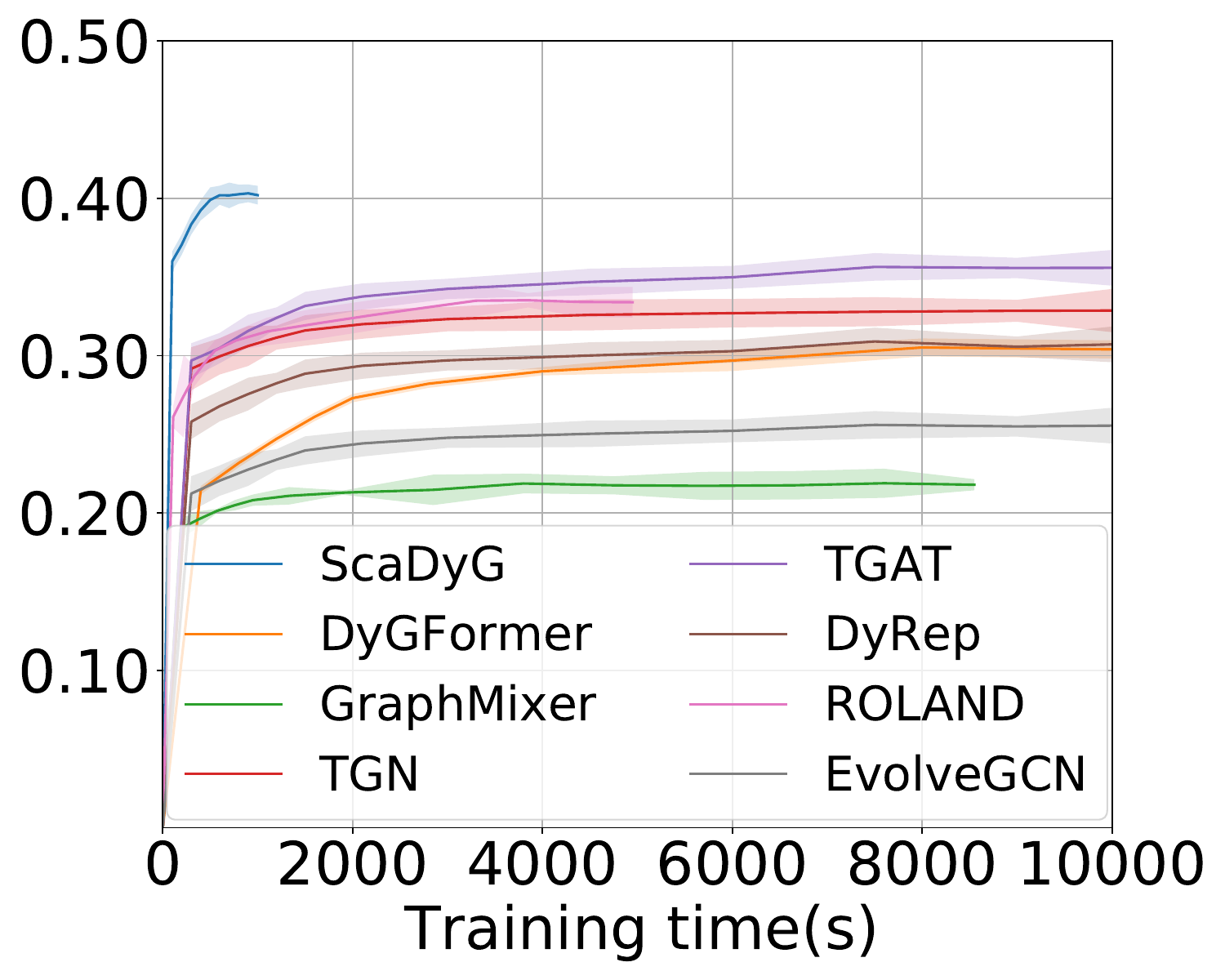}
        \caption{Training time on tgbl-reddit}
        \label{fig:subfig4}

    \end{subfigure}
\vspace{-0.2cm}
    \caption{Efficiency comparison of ScaDyG and baselines.}
    \vspace{-0.3cm}
    \label{fig:main}
    \label{fig_scability}

\end{figure*}

\subsection{Scalability performance  }

    To answer \textbf{Q2}, we first provide a theoretical analysis of algorithm complexity in Table~\ref{table_complexity}. 
    Subsequently, Fig.~\ref{fig_scability} compares the relative training times and convergence speeds of ScaDyG with other competitive baselines in link- and node-level evaluations. 
    Additionally, Table~\ref{efficiency_mooc}-\ref{efficiency_reddit} report the pre-processing time (Pre), average training time per epoch (E-train), average testing time per epoch (E-test), GPU memory usage for the same batch size (G-Mem), and the number of parameters (Param), offering key insights for practical deployment.
    
\noindent
\textbf{Complexity analysis.} 
    We first analyze the time and space complexity of ScaDyG. 
    In the pre-processing phase, ScaDyG propagates all edge features to nodes for $l$ hops, resulting in a time complexity of $O(lmf)$.
    During training, ScaDyG performs linear aggregation of historical steps of length $N$, followed by weight generation and feature transformation, both involving only linear transformations, leading to a time complexity of $O(nNLf^2)$.
    Regarding space complexity, ScaDyG needs to store model parameters and node features for a batch in GPU memory.
    Since we generate a parameter matrix for each node in a batch, the space complexity for storing these parameters is $O(bf^2)$. 
    Additionally, the space complexity for storing intermediate messages for $L$ steps is $O(bLf)$.
    In summary, the total space complexity is $O(bLf+bf^2)$. Discrete-based methods encounter OOM (Out-Of-Memory) issues on the StackOverflow dataset due to the spatial complexity being assoicated to the number of nodes. Conversely, continuous-based methods, face OOT (Out-Of-Time) issues on this dataset because their link prediction complexity is related to the number of edges.

\begin{table}[]
    \caption{Epoch and batch efficiency on MOOC dataset.}
\vspace{-0.3cm}
\resizebox{\linewidth}{!}{\begin{tabular}{cccccc}
\hline
Method & Pre & E-train  & E-eval &G-Mem & Param \\
\hline
EvolveGCN &- & 363.78s &97.44s & 4573M & 2578K  \\
ROLAND & -&191.21s  & 43.53s & 3372M & 383K \\
JODIE & -& 156s  & 60.13s & 1326M &311K   \\
DyRep & -& 2676.31s & 528.45s & 4600M & 1188K \\

TGAT  &- & 2413.27s & 575.53s & 3526M & 1052K \\

TGN & - & 4479.01s & 993.32s & 4188M & 1460K\\

GraphMixer &- & 1128.64s & 124.41s & 2958M & 643K\\

DyGFormer & -& 2152.91s &  337.34s & 1942M & 976K\\
\hline
ScaDyG & 21.13s  & 126.43s & 53.78s & 1136M  & 55K\\
\hline
\end{tabular}
}
\label{efficiency_mooc}
\vspace{-0.4cm}
\end{table}

\begin{table}[]
    \caption{Epoch and batch efficiency on tgbl-reddit dataset.}
\vspace{-0.3cm}
\resizebox{\linewidth}{!}{\begin{tabular}{cccccc}
\hline
Method &Pre & E-train  & E-eval &G-Mem & Param \\
\hline
EvolveGCN &- &991.96s &  1008.24s& 22450M& 1018K\\
ROLAND &- & 256.29s & 247.64s & 18970M & 527K \\
JODIE &- &1057s & 798.5s &14942M  & 132K\\
DyRep &- & 974.51s  & 1492.29s & 15556M      & 244K \\

TGAT &- & 1350.25s & 1441.36s & 15212M & 712K\\

TGN &- & 1272s & 1466.58s & 15784M & 947K \\

GraphMixer &- & 413.72s & 1215.76s & 14484M & 381K \\

DyGFormer &- &1966.34s  & 939.55s &  15200M & 1067K\\
\hline
ScaDyG & 14.36s & 50.78s & 49.68s & 6178M & 182K\\
\hline
\end{tabular}
}
\label{efficiency_reddit}
\vspace{-0.4cm}
\end{table}

\noindent
\textbf{Running efficiency.}
    Subsequently, we aim to provide a more comprehensive evaluation from the perspective of practical running efficiency. 
    Specifically, we observe that ScaDyG is the most efficient method and give the following key insights. 
    (1) Time cost:
    Although ScaDyG introduces pre-processing compared to baselines, this is negligible relative to the total training duration and is executed only once. 
    Furthermore, ScaDyG shows superior performance in both total training time and training time per epoch, thanks to two key designs: 
    (i) Feature propagation is handled during pre-processing, reducing training overhead;
    (ii) A computation-friendly neural architecture is simple yet effective with fewer parameters and faster convergence. 
    As shown in the complexity analysis in Table \ref{table_complexity}, ScaDyG's efficient training time complexity stems from its minimal dependence on the number of edges, with edge-related operations limited to straightforward feature matrix transformations.
    (2) Memory cost:
    ScaDyG requires fewer trainable parameters, reducing the storage needed for parameter gradients and model states, thereby lowering GPU memory usage.
    Notably, ScaDyG needs additional storage for propagated features within pre-process, which are stored in main memory, avoiding extra GPU memory consumption.
    Further analysis of baseline results is provided in Appendix~\ref{baseline_results}.

% Spatial complexity primarily consists of the number of model parameters and the size of training data. ScaDyG  employs mini-batch training, reducing the spatial overhead to a linear relationship with batch size. 
% Within the continuous-time methods, memory module-based approaches have even greater storage costs, as they need to maintain the state of all nodes. On the other hand, sequence model-based  methods,e.g. DyGFormer, only need to store the representations of nodes within a batch in GPU memory, resulting in lower space complexity.

% We observe that among the baseline methods, discrete-time approaches exhibit faster training times compared to continuous-time methods but incur higher storage overhead. This is because they avoid time-consuming sampling and neighborhood aggregation processes but require storing node representations for historical snapshots.  ScaDyG eschews the inefficient designs of existing methods, thus enjoying better scalability performace. 

\begin{table}
\small
\caption{Results of ablation experiments.}
\vspace{-0.2cm}
\resizebox{\linewidth}{!}{\begin{tabular}{c|c c c c  }
\hline
Datasets & ScaDyG & w/o T & w/o HN & w/o TE \\
\hline
UCI &0.534 $\pm$ 0.007 &0.485
$\pm$ 0.009 &0.521 $\pm$ 0.013 
 & 0.470 $\pm$ 0.007 \\
MOOC  & 0.931 $\pm$ 0.009& 0.473 $\pm$ 0.009& 0.785 $\pm$ 0.011&  0.794 $\pm$ 0.016 \\
BitcoinAlpha & 0.597 $\pm$ 0.019&0.561 $\pm$ 0.016 & 0.567 $\pm$ 0.013& 0.575 $\pm$ 0.010 \\
LastFM & 0.248 $\pm$ 0.012&0.225 $\pm$ 0.008 &0.229 $\pm$ 0.012 & 0.231 $\pm$ 0.011 \\
UNvote &0.725 ± 0.009 &0.675 $\pm$ 0.012 & 0.697 $\pm$ 0.006 & 0.652 $\pm$ 0.011  \\
% Flights &0.486 ± 0.009& 0.445 $\pm$ 0.008&0.457 $\pm$ 0.006 &0.403 $\pm$ 0.006  \\
Reddit-title &0.679 ± 0.013 & 0.462 $\pm$ 0.021& 0.579 $\pm$ 0.009 & 0.584 $\pm$ 0.006 \\
Enron &0.481 $\pm$ 0.009 & 0.381 $\pm$ 0.010 &0.413 $\pm$ 0.009 & 0.401 $\pm$ 0.005 \\
Stackoverflow &0.559 ± 0.012 &0.471 $\pm$ 0.014 & 0.501 $\pm$ 0.009 & 0.446 $\pm$ 0.008 \\
\hline
tgbl-trade & 0.631 ± 0.007 &0.375 $\pm$ 0.006 &0.390 $\pm$ 0.007 & 0.383 $\pm$ 0.019   \\
tgbl-genre & 0.403 ± 0.005 &0.294 $\pm$ 0.009 &0.357 $\pm$ 0.015 & 0.335 $\pm$ 0.012 \\
tgbl-reddit &0.402 $\pm$ 0.012  &0.303 $\pm$ 0.011& 0.356 $\pm$ 0.012 & 0.327 $\pm$ 0.010  \\
tgbl-token & 0.686 $\pm$ 0.019 & 0.578 $\pm$ 0.021& 0.639 $\pm$ 0.011 & 0.623  $\pm$ 0.014  \\
\hline
\end{tabular}
}
\label{ablation}
\vspace{-0.2cm}
\end{table}

\begin{figure}[]

	\centering
	\begin{minipage}{0.49\linewidth}
		\centering
		\includegraphics[width=0.95\linewidth]{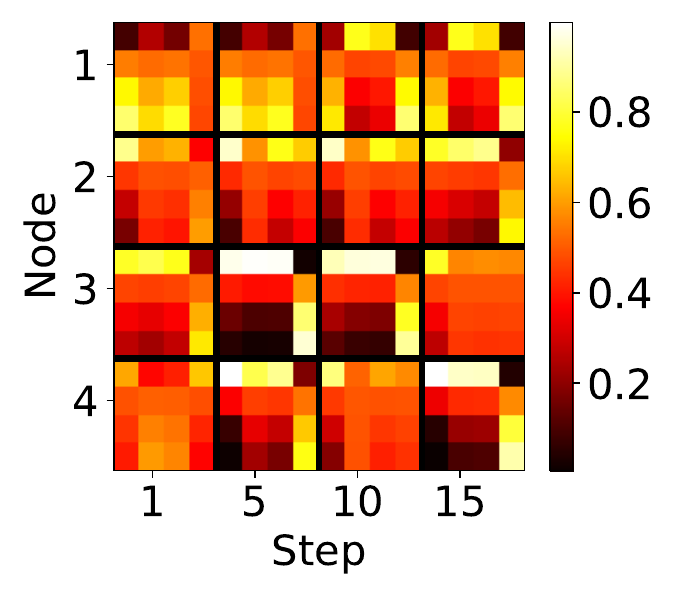}
           
		\subcaption{On MOOC dataset.}
		\label{chutian1}%文中引用该图片代号
    
	\end{minipage}
	%\qquad
	\begin{minipage}{0.49\linewidth}
		\centering
		\includegraphics[width=0.95\linewidth]{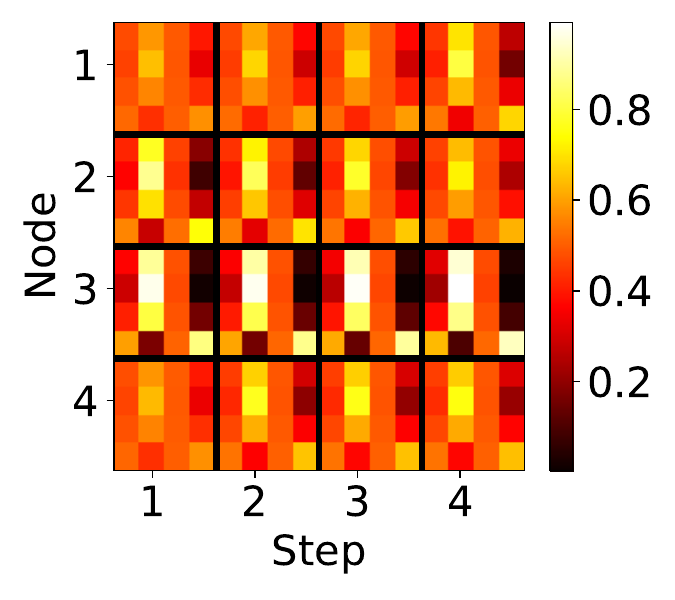}
           
            \subcaption{On tgbl-trade dataset.}
		\label{chutian2}%文中引用该图片代号
	\end{minipage}
\vspace{-0.2cm}
	\caption{ Visualization of transformation matrix.}
 \vspace{-0.2cm}
	\label{fig_visualize}

\end{figure}

\subsection{Interpretability Analysis}
\label{Interpretability}

    To answer \textbf{Q3}, we first conduct an ablation study to verify three key components of our proposed ScaDyG, followed by the visualization analysis of the weights of the transformation matrix generated by the hypernetwork in inference stage. 
    Specifically, in the ablation study, "w/o T" represents ScaDyG without temporal encoding, where the sum of the historical step node features through the hypernetwork is used as the final node embedding.  
    "w/o HN" represents ScaDyG without Hypernetwork-driven message aggregation, using $\mathbf{X}_t$ directly for downstream prediction. \textcolor{black}{
    "w/o TE" involves replacing the combination of exponential functions with a traditional exponential time modeling \cite{zuo2018:embedding,wen2022:trend}.  We repeat the single exponential function $d_e$ times as the temporal encoding vector. }
    For visualization analysis, we display the top-left 4x4 submatrix of the transformation matrix of different nodes at different steps.

% To further illustrate the effect of the Hypernetwork-dirven Message Aggregation, we visualize the weights of the transformation matrix generated by the hypernetwork on the MOOC and tgbl-trade datasets in Figure \ref{fig_visualize}. For brevity, we only display the top-left 4x4 submatrix of the weights matrix.

\noindent
\textbf{Result of ablation experiment.} 
    From Table~\ref{ablation}, we have the following observations: 
    (1) The most significant performance degradation occurs in the "w/o TE" model, underscoring the importance of dynamic dependencies in historical interactions;
    (2) \textcolor{black}{The removal of either Hypernetwork-driven Message Aggregation (w/o HN) or the combination of exponential functions (w/o TE) leads to substantial performance drops, indicating that the hypernetwork is crucial for dynamic encoding, and the proposed time encoding outperforms traditional exponential time modeling.
    The absence of either compromises the node-wise dynamic temporal modeling capability, resulting in insufficient representation of temporal dependencies.}

\noindent
 \textbf{Visualization Analysis.} 
    Based on Fig.~\ref{fig_visualize}, we observe that different nodes exhibit distinct weight patterns, and the weights of the same node gradually evolve over different steps. 
    Specifically, in the MOOC dataset, the weights vary more significantly between nodes and even within the same node (e.g., nodes 3 and 4) across various steps.
    This variation indicates diverse patterns between nodes and significant changes in the graph structure between steps. 
    In contrast, in the tgbl-trade dataset, the patterns are more similar between different nodes. 
    Within the same step, the weights show a clearer gradual change, reflecting the temporal evolution trend in the dynamic graph. 
    These results further validate the effectiveness of Hypernetwork-driven Message Aggregation in dynamically modeling node-wise temporal patterns.

\subsection{Hyperparameter Sensitivity}
\label{param}
\begin{figure}
    \centering

    \begin{subfigure}{0.23\textwidth}
        \centering
        \includegraphics[width=\textwidth]{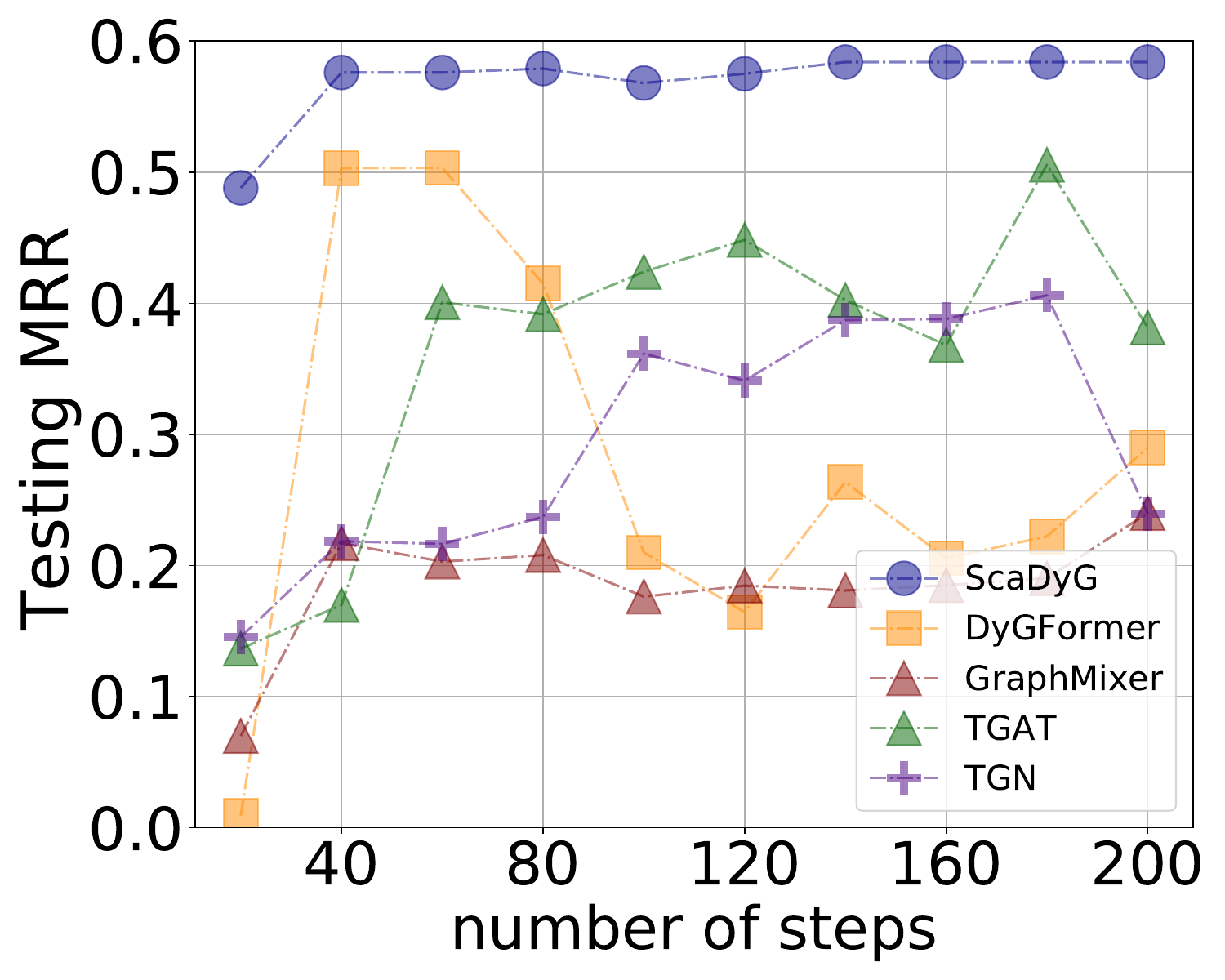}
        \caption{ Link prediction on BitcoinAlpha}
        \label{fig:subfig1}
    \end{subfigure}
    \hfill
    \begin{subfigure}{0.23\textwidth}
        \centering
        \includegraphics[width=\textwidth]{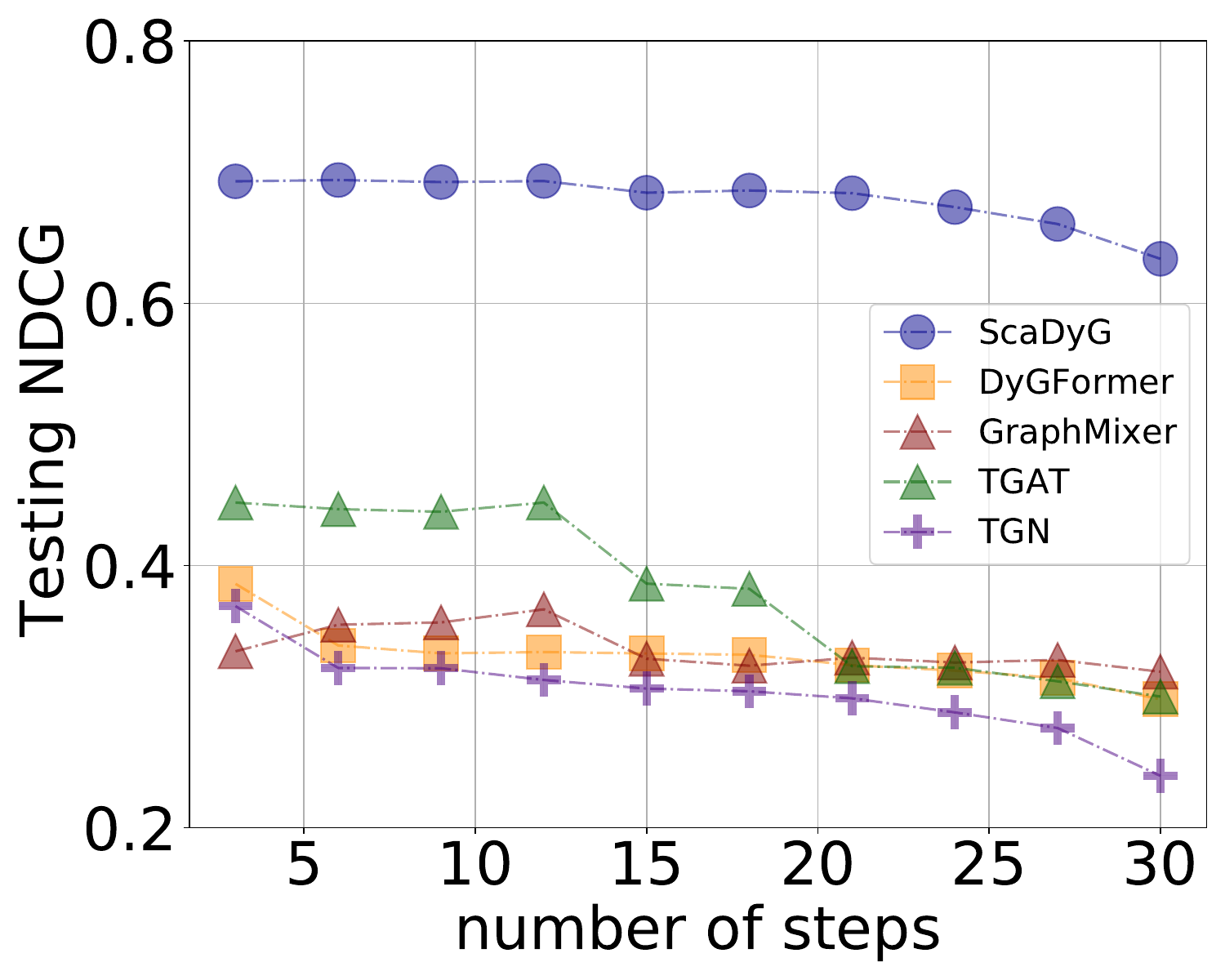}
        \caption{ Node affinity prediction on genre}
        \label{fig:subfig2}
    \end{subfigure}
    \hfill
    \begin{subfigure}{0.23\textwidth}

        \centering
        \includegraphics[width=\textwidth]{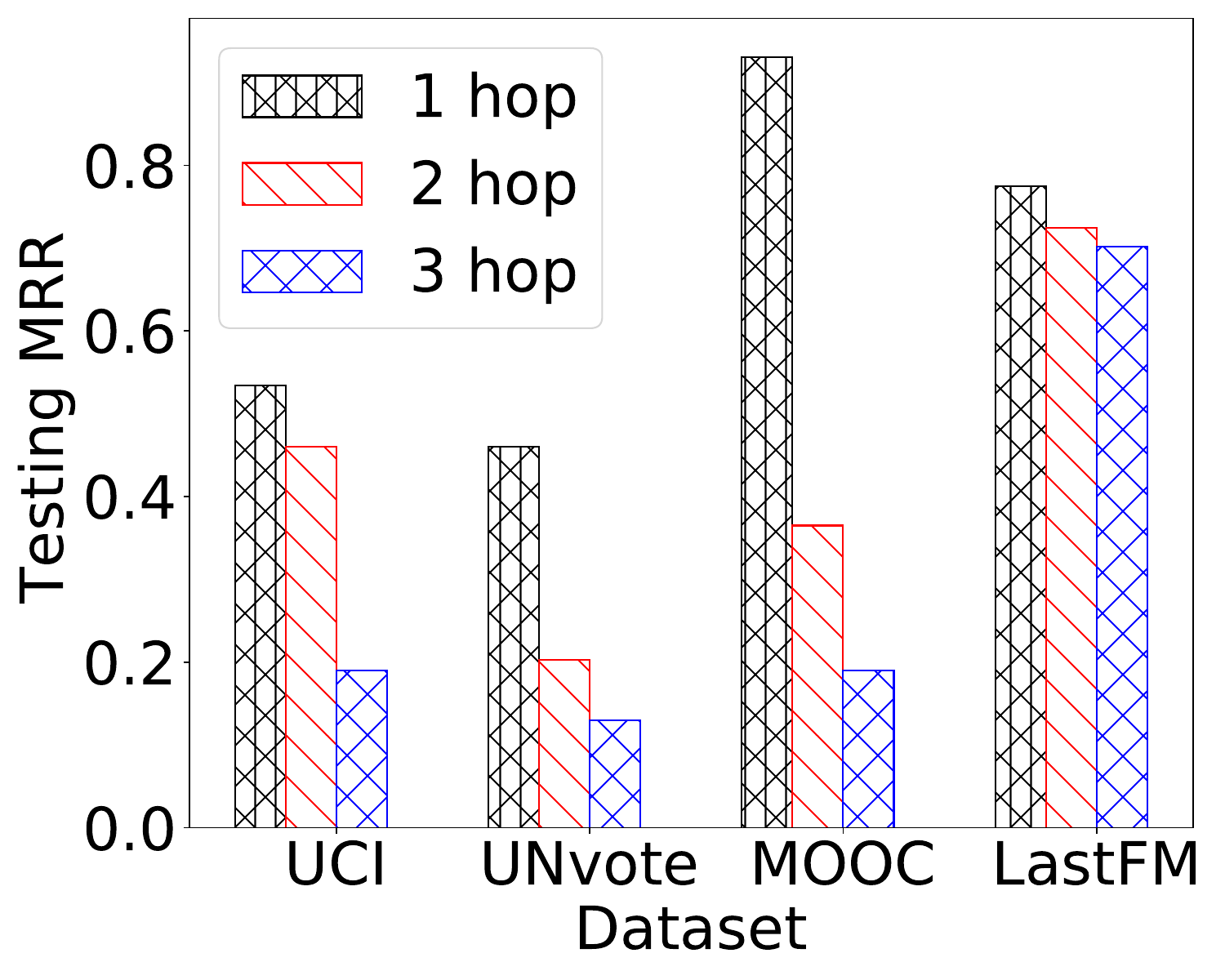}
        \caption{ Link prediction }
        \label{fig:subfig3}
    \end{subfigure}
    \hfill
    \begin{subfigure}{0.23\textwidth}
        \centering
        \includegraphics[width=\textwidth]{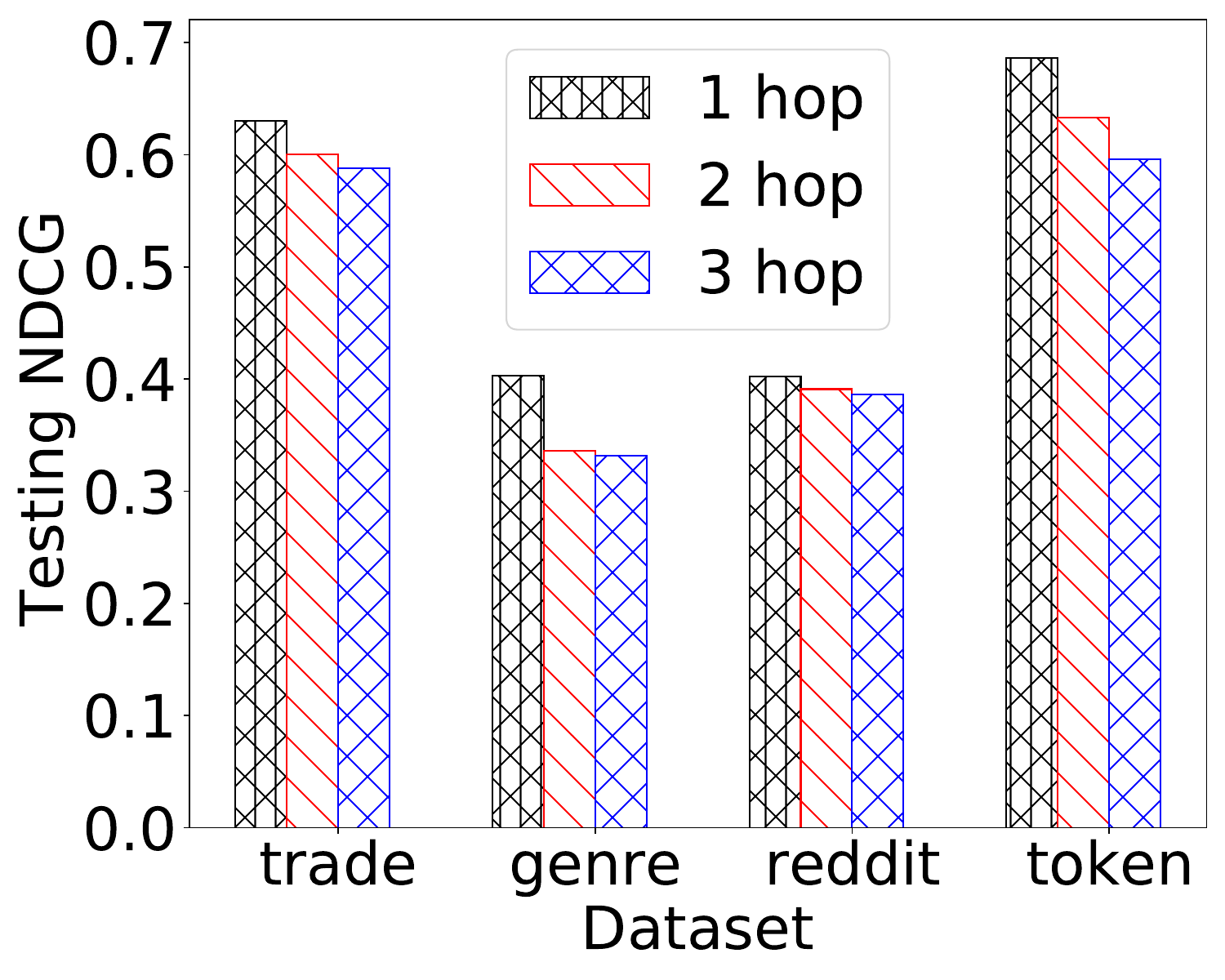}
        \caption{ Node affinity prediction }
        \label{fig:subfig4}
    \end{subfigure}
\vspace{-0.3cm}
    \caption{Results of hyperparameter study.}
    \label{fig:main}
    \label{fig_hyperparameter}
\vspace{-0.6cm}
\end{figure}

    To answer \textbf{Q4}, we explore two key performance parameters: historical steps and neighborhood hops. 
    Specifically, historical steps determine the length of the information window, with more steps capturing long-range dependencies and fewer steps focusing on short-range ones.
    Neighborhood hops define the range of neighborhood information used, with single hops focusing on direct interactions and multiple hops integrating broader propagation. 
    The experimental results are shown in Figure.~\ref{fig_hyperparameter}(a)-(b) and (c)-(d).

\noindent
\textbf{Historical steps.}
    Experimental results show that ScaDyG is less sensitive to the number of steps compared to baselines.
    Specifically, in the link prediction task on the BitcoinAlpha, ScaDyG's optimal performance is observed with parameters in the 160-200 range, emphasizing the importance of long-range dependencies in this dataset. 
    In contrast, the optimal parameters for baselines vary inconsistently, suggesting that their performances are affected by the randomness of neighborhood sampling. 
    For the node affinity prediction task on the tgbl-genre, most methods perform best with a time step of 3, with MRR decreasing as the number of time steps increases. 
    This decline may be attributed to the presence of noise in long-range temporal information within the tgbl-trade.

\noindent
\textbf{Neighborhood hops.} 
    According to the experimental results shown in Fig.~\ref{fig_hyperparameter}(c)-(d), we observe that for both link prediction and node affinity prediction, performance is optimal at 1-hop and significantly declines with multiple hops. 
    This suggests that incorporating multi-hop information can be detrimental to TG tasks, contrasting with results on simple static graphs \cite{frasca:2020sign, li2024_atp}.
    One possible explanation is that multi-hop interactions on DGs often introduce much noise, potentially degrading the model's predictive performance.

\section{Conclusion}
In this study, we propose ScaDyG, an efficient and effective framework for learning on large-scale dynamic graphs. By reformulating the message passing process of dynamic GNNs within a decoupled framework, ScaDyG achieves superior time and space efficiency compared to existing methods, making it particularly well-suited for modeling million-level dynamic graphs. Unlike previous approaches that rely on sequence-based models for temporal modeling, ScaDyG employs a simple yet powerful time encoding scheme based on a dynamic fusion of exponential functions. To extend temporal modeling to node-level granularity, the preprocessed temporal messages are aggregated through a hypernetwork-driven transformation. Interestingly, we observe diverse temporal patterns in the generated weights.  The designs not only reduce the number of parameters and training time but also enable ScaDyG to deliver competitive results across a diverse range of dynamic graphs.  For future work, we will explore broader applications of scalable frameworks on a large scale, e.g., billion-level dynamic graph learning.

%It utilizes a decoupled architecture, where the historical information is segmented to multiple time steps, the node state at each time step is achieved through efficient dynamic feature propagation. To utilize complete temporal information encoding, an exponential-function-based temporal encoding is designed to integrate  node state at multi time steps.  We  conduct experiments on both link and node levels across a wide range of datasets, surpassing both discrete and continuous-time benchmark methods.  For future work, we will explore broader applications of scalable frameworks in dynamic graph learning.

\bibliographystyle{ACM-Reference-Format}
\bibliography{sample-base}

\newpage
\clearpage

\appendix

% \begin{align}
%      \mathbf{m}_v^{t_i} &= \mathbf{h_v} \odot T_e(t_i-t') \\
%     &= [x_1e^{\gamma_1t}, x_2e^{\gamma_2t},..,x_de^{\gamma_dt}]
% \end{align}

\section{Proof of proposition \ref{prop1}}
\label{proof}

Assuming the  initial edge feature of  $(u,v,t')$ is $ \mathbf{x}_{e_v} = [x_1,x_2,...,x_d ] $, the time feature  is $[e^{\gamma_1t}, e^{\gamma_2t},...,e^{\gamma_dt}]$. let $\Delta t = \Delta t_1 +\Delta t_2$,
 $\mathbf{x}_{e_v}\odot T_e(\Delta t_1)\odot T_e(\Delta t_2)\mathbf{W}_1$ is , 
\begin{equation}
    \begin{split}
        & \mathbf{x}_{e_v} \odot T_e(\Delta t_1) \odot T_e(\Delta t_2)\mathbf{W}_1 \\
        &= [x_1e^{\gamma_1 \Delta t_1}e^{\gamma_1 \Delta t_2}, x_2e^{\gamma_2 \Delta t_1}e^{\gamma_2 \Delta t_2}, \ldots, x_de^{\gamma_d \Delta t_1}e^{\gamma_d \Delta t_2}] \mathbf{W}_1 \\
        &=  [x_1e^{\gamma_1 (\Delta t)}, x_2e^{\gamma_2 (\Delta t)}, \ldots, x_de^{\gamma_d (\Delta t)}]\mathbf{W}_1.
    \end{split}
\end{equation}

\noindent
Let $\mathbf{W}_1 \in \mathbb{R}^{d\times d }$  be expressed as:

\begin{equation}
\mathbf{W_1}=
\begin{bmatrix}
w_{11} & w_{12} & \cdots & w_{1d} \\
w_{21} & w_{22} & \cdots & w_{2d} \\
\vdots & \vdots & \ddots & \vdots \\
w_{d1} & w_{d2} & \cdots & w_{dd}
\end{bmatrix}
\end{equation}
thus,

\begin{equation}
\begin{split}
   &\mathbf{x}_{e_v}\odot T_e(\Delta t_1) \odot T_e(\Delta t_2)\mathbf{W}_1  \\
   &= \begin{bmatrix}
   w_{11}x_1e^{\gamma_1 \Delta t} + w_{12}x_2e^{\gamma_2 \Delta t} + \cdots + w_{1d}x_de^{\gamma_d \Delta t} \\
   w_{21}x_1e^{\gamma_1 \Delta t} + w_{22}x_2e^{\gamma_2 \Delta t} + \cdots + w_{2d}x_de^{\gamma_d \Delta t} \\
   \vdots \\
   w_{d1}x_1e^{\gamma_1 \Delta t} + w_{d2}x_2e^{\gamma_2 \Delta t} + \cdots + w_{dd}x_de^{\gamma_d \Delta t}
   \end{bmatrix}^T
\end{split}
\end{equation}

\noindent
On the other hand,

Let $\mathbf{W} \in \mathbb{R}^{d\times d }$  be expressed as:

\begin{equation}
\mathbf{W}=
\begin{bmatrix}
z_{11} & z_{12} & \cdots & z_{1d} \\
    z_{21} & z_{22} & \cdots & z_{2d} \\
    \vdots & \vdots & \ddots & \vdots \\
    z_{d1} & z_{d2} & \cdots & z_{dd}
\end{bmatrix},
\end{equation}
then,
\begin{equation}
    \begin{split}
    &\mathbf{x}_{e_v} \kappa (\Delta t_1+\Delta t_2) )\mathbf{W}  \\ 
    &= \mathbf{x}_{e_v}\kappa (\Delta t) )\mathbf{W} \\ 
    &= \mathbf{x}_{e_v} (a_1e^{\gamma_1 \Delta t}+a_2e^{\gamma_2 \Delta t}+\cdots+a_de^{\gamma_ \Delta t})\mathbf{W} \\ 
    &= \sum_{i=1}^d a_ie^{\gamma_i (\Delta t)}  \begin{bmatrix}
    x_{1} \\
    x_{2}  \\
    \vdots \\
    x_{d} 
    \end{bmatrix}^T 
    \begin{bmatrix}
    z_{11} & z_{12} & \cdots & z_{1d} \\
    z_{21} & z_{22} & \cdots & z_{2d} \\
    \vdots & \vdots & \ddots & \vdots \\
    z_{d1} & z_{d2} & \cdots & z_{dd}
    \end{bmatrix} 
     \\ 
    &= \begin{bmatrix}
    z_{11}x_{1} + z_{12}x_{2} + \cdots + z_{1d}x_{d} \\
    z_{21}x_{1} + z_{22}x_{2} + \cdots + z_{2d}x_{d} \\
    \vdots \\
    z_{d1}x_{1} + z_{d2}x_{2} + \cdots + z_{dd}x_{d}
    \end{bmatrix}^T
    \sum_{i=1}^d a_ie^{\gamma_i (\Delta t)} 
     \label{eq12}
\end{split}
\end{equation}
% \begin{align}
%     \mathbf{W_2} \mathbf{h}_v \theta (\Delta t_1+\Delta t_2) ) = \mathbf{W}_2 \mathbf{h}_v \theta (\Delta t) ) = \mathbf{W_2h}_v (a_1e^{\gamma_1 \Delta t}+a_2e^{\gamma_2 \Delta t}+\cdots+a_ne^{\gamma_1 \Delta t}) \\
%     &= \begin{bmatrix}
% l_{11} & l_{12} & \cdots & l_{1d} \\
% l_{21} & l_{22} & \cdots & l_{2d} \\
% \vdots & \vdots & \ddots & \vdots \\
% l_{d1} & l_{d2} & \cdots & l_{dd}
% \end{bmatrix} \begin{bmatrix}
% x_{1} \\
% x_{2}  \\
% \vdots \\
% x_{d} 
% \end{bmatrix} 
% (a_1e^{\gamma_1 (\Delta t)}+a_2e^{\gamma_2 (\Delta t)}+\cdots+a_ne^{\gamma_d (\Delta t)})\\
% &= \begin{bmatrix}
% l_{11}x_{1} + l_{12}x_{2} + \cdots + l_{1d}x_{d} \\
%  l_{21}x_{1} + l_{22}x_{2} + \cdots + l_{2d}x_{d} \\
% \vdots \\
% l_{d1}x_{1} + l_{d2}x_{2} + \cdots + l_{dd}x_{d}
% \end{bmatrix}
% (a_1e^{\gamma_1 (\Delta t)}+a_2e^{\gamma_2 (\Delta t)}+\cdots+a_ne^{\gamma_d (\Delta t)})
% \label{eq12}
% \end{align}

By assigning $w_{ij} = z_{ij}(a_1e^{(\gamma_1-\gamma_j) \Delta t}+a_2e^{(\gamma_2-\gamma_j) \Delta t}+\cdots+a_ne^{(\gamma_d-\gamma_j) \Delta t})$, $i,j = 1,2,...,d$ the proposition  is proved.

% \subsection{Proof of proposition \ref{prop2}}

% Assume, the weights $[a_1,a_2,...,a_{d_e}]$ are fixed values during the training, then  $(a_1e^{(\gamma_1-\gamma_j) \Delta t}+a_2e^{(\gamma_2-\gamma_j) \Delta t}+\cdots+a_ne^{(\gamma_d-\gamma_j) \Delta t})$ is a fixed value. We  suppose the value is $a$. Through equation \ref{eq12}, we can see there is a multiplicative relationship between $w_{ij}$ and $l_{ij}$: 
% \begin{equation}
%     z_{i,j}= \frac{w_{i,j} e^{\gamma_j \Delta t}}{a},
% \end{equation}
%  which contradicts the premise that $\mathbf{W}$ and $\mathbf{W}_1$ are learnable parameter matrices (which means that there is no predetermined multiplicative  relationship between the parameters of two matrices).
%= w_{i,j} \frac{e^{\gamma j}{k}$

\section{Additional experiments}

\subsection{Additional experimental settings}
\label{detail}
\noindent
\textbf{Implementation settings.} The hyperparameters of ScaDyG mainly include the learning rate, the parameter $\{\gamma_i\}_{i=1}^{d_{e}}$ for the combination of exponentials and the number of historical steps. The learning rate is searched from \{$10^{-5}$,$10^{-1}$\}, and $n$ is searched from 1 to the total time steps in the specific dataset. \textcolor{black}{ The parameters of the exponential functions in time encoding,i.e. $\{\gamma_i\}_{i=1}^{d_{e}}$, can be any combination of different values.  We set them as a uniformly decreasing sequence for convenience, i.e., $\gamma_i = \gamma_0 - id$, $d = (\gamma_{d_0} -\gamma_{d_e}) / d_e$ and  $\gamma_{d_e} = 10^{-1} \gamma_0$. Since different datasets have varying time ranges, we found that setting $\gamma_0$
  to be around the inverse of the dataset's time range works best. }     For datasets that do not contain edge features, we follow \cite{you2022:roland} to generate a one-dimensional feature based on the edge timestamps. For datasets with only one-dimensional edge features, we replicate the edge features into 8-dimensional vectors to facilitate dynamic feature encoding. For dynamic link prediction tasks,  the methods are evaluated under the ranking setting, as it can avoid the bias of easy negatives compared to binary classification \cite{huang2024:temporal}. However, we also provide the results under binary classification in Appendix \ref{add_results} for reference. For baselines, we use the library proposed by \cite{yu2023towards} for the implementation of the five continuous-based methods, i.e, DyGFormer, GraphMixer, TGAT, TGN, JODIE, and DyRep. ROLAND and EvolveGCN are adopted from their official implementation. The same node and edge feature dimensions are set equal to ScaDyG. All experiments are conducted on a Linux server with an Intel Xeon Gold 5218R CPU and an NVIDIA  RTX 3090 with 24GB memory. 

\noindent
\textbf{Evaluation metrics.} For dynamic link prediction task,  we employ a multi-layer perceptron to predict the presence of an edge between two nodes, using the concatenated embeddings of these nodes as input. The methods are evaluated under the ranking setting, where each positive sample is compared against 100 negative samples obtained through random sampling (with collision check). Compared to the binary classification, the ranking setting can avoid the bias of easy negatives, thus making the results more reliable \cite{huang2024:temporal}. MRR (Mean Reciprocal Rank) is selected as the metric to evaluate the ranking quality. For node affinity prediction, we use a multi-layer perceptron to predict the affinity vector based on the node representation, and we follow \cite{huang2024:temporal} use NDCG (Normalized Discounted Cumulative Gain) as the evaluation metric.

\noindent
\textbf{Additional background for tasks.} Most existing methods focus on link prediction tasks \cite{wen2022:trend,cong2023:we,you2022:roland,xu2020:inductive,rossi2020:temporal}, with only a few considering node classification \cite{lu2019:temporal,yu2023towards,pareja2020:evolvegcn}. This is mainly due to the lack of labeled data for nodes in most temporal graph datasets. The few datasets with node labels also suffer from two significant limitations: 1) Their labels are binary (0 or 1), which does not allow for a comprehensive evaluation of node-level performance. 2) Their labels do not change over time, failing to assess the model's ability to capture temporal dynamics. Therefore, we did not use node classification tasks but instead adopted a newly introduced node affinity prediction task. This task's datasets avoid the mentioned limitations and are of a larger scale. We provide a detailed introduction to these datasets in  Appendix \ref{datasets}.

\noindent
\subsection{Description of datasets} 
\label{datasets}
\textbf{Dataset for dynamic link prediction.} In this section, we give a brief introduction to the datasets. For dynamic link prediction, we conducted experiments on 8 public datasets, including UCI, MOOC, UNvote, BitcoinAlpha, Stackoverflow, Reddit-title, and Enron.    These datasets are uniformly segmented into time steps based on predefined intervals. The time steps are split chronologically with a ratio of 70\%/15\%15\% for training, validation, and testing.
\begin{itemize}
\item[$\bullet$]\textbf{UCI} \cite{poursafaei:2022towards, panzarasa2009:patterns}  is a social network, providing records of interaction among students from the University of California at Irvine. Each interaction $(u,v,t)$ characterizes a message between two users at a timestamp down to the second.

\item[$\bullet$]  \textbf{MOOC}  \cite{poursafaei:2022towards,kumar:predicting} constitutes a network of student interactions derived from components of online courses, including problem sets and video materials. Each interaction $(u,v,t)$ represents a student engaging with a piece of content, characterized by four distinct attributes.

\item[$\bullet$] \textbf{LastFM} \cite{poursafaei:2022towards,kumar:predicting} is a dataset from the online music platform Last.fm, each interaction $(u,v,t)$ in the dataset represents user $u$ listening to a song $v$ at time $t$. The dataset encompasses the listening activities of 1000 users with respect to the top 1000 songs over a timeframe of one month.

% \item[$\bullet$] \textbf{Wikipedia} \cite{poursafaei:2022towards,kumar2019:predicting} is a dataset collected from Wikipedia, where each interaction $(u,v,t)$ denotes a user $u$ editing a page $v$ at time $t$. Initial features of the pages are encoded from the edited text, while initial features of users are defined as the sum of the features of the pages they edited.

\item[$\bullet$] \textbf{Bitcoinalpha} \cite{kumar:2016edge,kumar:2018rev2} represents a trust-based network of bitcoin users engaged in transactions via the Alpha platform. Each interaction $(u,v,t)$ denotes a rating from one user to another.

\item[$\bullet$] \textbf{UNvote} \cite{poursafaei:2022towards} tracks the roll-call votes in the United Nations General Assembly. Each interaction represents a joint voting behavior between two nations.   The weight of the link between them corresponds to the number of joint votes in a year.

\item[$\bullet$] \textbf{Reddit-title} \cite{kumar2018:community} is constructed based on the hyperlink connections between subreddits on the Reddit platform. Each temporal edge represents a hyperlink in the title of a post from one subreddit to another. The timestamp associated with the edge is the creation time of the post.

\item[$\bullet$] \textbf{Enron} \cite{poursafaei:2022towards} encompasses email communications among employees within the Enron company. Each interaction $(u,v,t)$ represents an email sent from $u$ to $v$.

% \item[$\bullet$] \textbf{Flights} \cite{huang2024:temporal} comprises a crowdsourced  international flight network  from 2019 to 2022. Airports are represented as nodes, while the edges represent flights between airports on a given day.  Each interaction denotes a flight between two airports on a given day. 
\item[$\bullet$] \textbf{Stackoverflow} \cite{paranjape:2017motifs} captures user interactions within the Stack Overflow community, where nodes correspond to individual users and each interaction $(u,v,t)$ indicates the act of one u providing an answer to v's inquiry at $t$.
\end{itemize}

\noindent
\textbf{Datasets for dynamic node affinity prediction.}  For dynamic node affinity prediction task, we adopt 4 datasets provided by \cite{huang2024:temporal}, encompassing tgbl-genre, tgbl-trade, tgbl-reddit and tgbl-token.  We chose these four datasets because the node affinity labels in the datasets are multi-dimensional and associated with time steps, allowing a more comprehensive evaluation than binary or static labels \cite{huang2024:temporal}. The node affinity labels in the datasets are associated with time steps. Accordingly, we divide the training, validation, and test set according to the time steps in chronological order at a ratio of 70\%/15\%15\%.  We give a brief introduction of the dataset as follows.

\begin{itemize}
 
\item[$\bullet$] \textbf{tgbl-trade} \cite{huang2024:temporal}  records the international agriculture trade network among United Nations member countries from 1986 to 2016. In this network, each temporal edge $(u,v,t)$ represents a trading relation between country $u$ to another country $v$. The edge value signifies the total value of agricultural products traded in one year. The node affinity represents the percentage distribution of annual trade products from a specific country to other countries. 

\item[$\bullet$] \textbf{tgbl-genre} \cite{huang2024:temporal}   represents a user-item bipartite network capturing interactions between users and the music genres they prefer. Each temporal edge $(u,v,t)$  delineates a user $u$ has listened to a song of a specific genre $v$ at a particular time $t$. The edge weight signifies the degree of affiliation of the song to this genre, measured in percentages. For a user node, the property is the frequency of interactions between the user and music across all genres.

\item[$\bullet$] \textbf{tgbl-reddit} \cite{huang2024:temporal}   represents the interaction between users and subreddits in the Reddit platform. Each interaction $(u,v,t)$ represents the user made a post on the subreddit. The property of a user node delineates the frequency of its interactions with various subreddits in a week.  

\item[$\bullet$] \textbf{tgbl-token} \cite{huang2024:temporal}   is a network that maps user interactions with cryptocurrency tokens. A temporal edge between a user node and a token node signifies a transaction made by the user to acquire that specific token, and the weight of the edge represents the quantity of the token transferred in that transaction. The node affinity is the frequency of interactions that a user has with various types of cryptocurrency tokens over a period of one week.

\end{itemize}

\subsection{Description of baselines}
\label{baselines}

In this section, we give a brief introduction to the baselines used in comparison experiments.  
\begin{itemize}
\item[$\bullet$]\textbf{DyGFormer} \cite{yu2023towards}  introduces a  Transformer-based architecture for dynamic graph learning that leverages historical interactions through neighbor co-occurrence encodings and a patching technique for long-sequence processing.

\item[$\bullet$]  \textbf{GraphMixer}  \cite{poursafaei:2022towards,kumar2019:predicting} aims to reduce the complexity of temporal graph modeling with a straightforward yet effective architecture. It leverages MLP-based link encoders, mean-pooling for node information, and an MLP-based link classifier. The timestamps are projected by a static cosine-based encoding function.

\item[$\bullet$] \textbf{TGAT} \cite{xu2020:inductive} synthesizes temporal-topological neighborhood features and time-feature interactions through self-attention and a novel time encoding to generate dynamic node embeddings for temporal graphs.

\item[$\bullet$] \textbf{TGN} \cite{rossi2020:temporal} utilized a memory module to the dynamic state involving a node, which is updated on observing a new interaction. The current embedding is then obtained by aggregating the state and the messages received from neighboring nodes.

\item[$\bullet$] \textbf{DyRep} \cite{trivedi2019:dyrep} leverages two  processes, namely the communication and association.It employs a time-scale adaptive multivariate point process model to capture the evolution of dynamic graphs.

\item[$\bullet$] \textbf{ROLAND} \cite{you2022:roland} aims to repurpose state-of-the-art static GNN architectures. It treats node embeddings as evolving states that are hierarchically updated in a recurrent fashion over time. Furthermore, ROLAND features a live-update setting where predictions are made continuously, and the model is updated incrementally. It propose three variants namely ROLAND-Moving Average, ROLAND-MLP and ROLAND-GRU. We select the ROLAND-GRU as our baseline because its best reported performance.

\item[$\bullet$] \textbf{EvolveGCN} \cite{pareja2020:evolvegcn}  adapts graph convolutional network (GCN) models to the temporal dimension without depending on static node embeddings. It employs RNNs to evolve the GCN parameters dynamically to future time steps.

\item[$\bullet$] \textbf{JODIE} \cite{kumar2019:predicting} \textcolor{black}{ learns both static and dynamic user embeddings for users-items interaction graphs, representing their long-term and time-varying properties, respectively. It leverages coupled recurrent neural networks (RNNs) to update embeddings based on user-item interaction.}
\end{itemize}

\subsection{Additional experimental results}
\label{add_results}
\noindent
\textbf{Performance on binary classification.} \textcolor{black}{
Besides the ranking setting, we present comparative results for binary classification, which predicts which of two candidate nodes a given node will form an edge at a specific timestamp. We use AP and AUC as the evaluation metrics. The results are shown in table \ref{link_pred_ap} and \ref{link_pred_auc}, which show that ScaDyG still achieves the best performance. Additionally, on Bitcoinalpha and Reddit-title, most methods yield similar outcomes, making MRR a better metric for distinguishing between them.}

\begin{figure}
    \centering
    \includegraphics[width=0.35\textwidth]{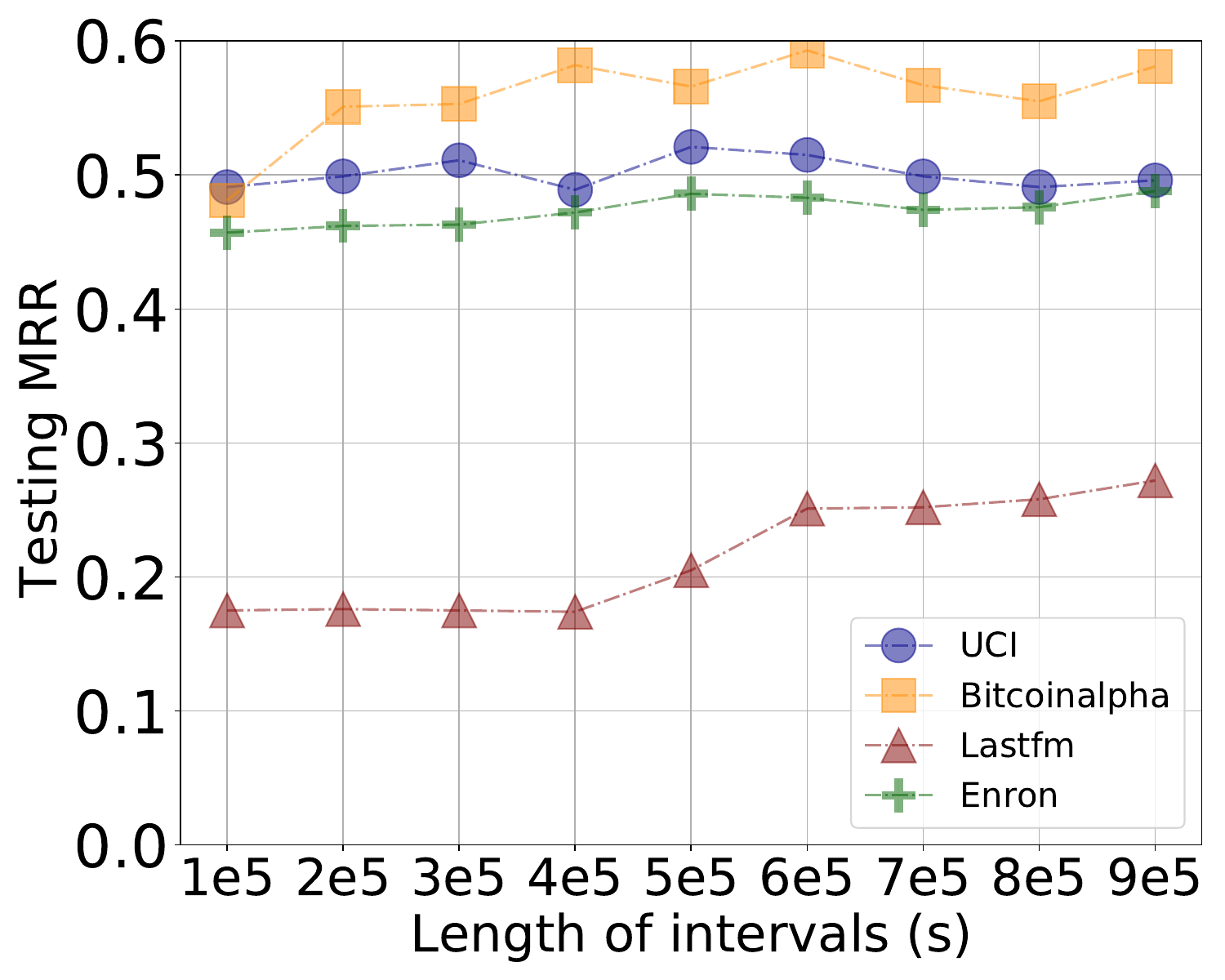} 
    \caption{Hyperparameter analysis on intervals.}
    \label{fig:hyperparameter_interval}
\end{figure}

\noindent
\textbf{Hyperparameter analysis on intervals.}
\textcolor{black}{ We conduct experiments to evaluate ScaDyG's behavior under different intervals between steps (Note that steps = time range/interval). The experiments are conducted only in link prediction, as the intervals for node affinity prediction are predetermined by the dataset.  
As shown in figure \ref{fig:hyperparameter_interval},  MRR across all datasets does not significantly decrease with increasing step intervals, indicating ScaDyG effectively preserves information between steps. On the Lastfm and Bitcoinalpha datasets, smaller intervals result in a drop in MRR, mainly because of the limitation of the information window, given the large number of steps.  Moreover, extremely small intervals are not preferred practically due to scalability considerations.}
% Given the significant variation in time spans across datasets, we present the results for each dataset separately.

\begin{table}[]
    \caption{Performance on binary classification setting (AP).}
\vspace{-0.2cm}
\resizebox{\linewidth}{!}{\begin{tabular}{ccccc}
\hline
Method  & MOOC  & Bitcoinalpha &Reddit-title &UCI \\
\hline
EvolveGCN  &0.728 ± 0.016   & 0.959 $\pm$ 0.003& 0.944 $\pm$ 0.003 &  0.808 $\pm$ 0.006 \\
ROLAND & 0.874 ± 0.016 & 0.983 $\pm$ 0.002 & 0.980 $\pm$ 0.003   & 0.894 $\pm$ 0.005 \\
JODIE & 0.805 ± 0.026  & 0.984 $\pm$ 0.001 & 0.991 $\pm$ 0.001  & 0.894 $\pm$0.010 \\
DyRep & 0.820 ±0.006  & 0.901 $\pm$ 0.001 & 0.984 $\pm$ 0.002 & 0.889 $\pm$ 0.003 \\

TGAT   & 0.856 ± 0.002  & 0.991 $\pm$ 0.001 & 0.990 $\pm$ 0.001 & 0.799 $\pm$ 0.009 \\

TGN  & 0.890 ± 0.018 & 0.992 $\pm$ 0.002  & 0.989 $\pm$ 0.001  & 0.929 $\pm$ 0.015 \\

GraphMixer  & 0.828 ± 0.003  & \underline{0.993} $\pm$ 0.001 & \underline{0.991} $\pm$ 0.001 & 0.931 $\pm$ 0.007\\

DyGFormer &\underline{0.873} ± 0.005 & \underline{0.993} ± 0.001 & \underline{0.991} $\pm$ 0.001 & \underline{0.953} $\pm$ 0.002 \\
\hline
ScaDyG & \textbf{0.901} $\pm$ 0.012 & \textbf{0.996} $\pm$ 0.001 & \textbf{0.992} $\pm$ 0.002 & \textbf{0.956} $\pm$ 0.002\\
\hline
\end{tabular}
}
\label{link_pred_ap}

\end{table}

\begin{table}[]
    \caption{Performance on binary classification setting (AUC).}
\vspace{-0.2cm}
\resizebox{\linewidth}{!}{\begin{tabular}{ccccc}
\hline
Method  & MOOC  & Bitcoinalpha &Reddit-title &UCI \\
\hline
EvolveGCN  & 0.808 ± 0.005 &0.955 $\pm$ 0.001 & 0.947 $\pm$ 0.003&  0.802 $\pm$ 0.008 \\
ROLAND & 0.854 ± 0.013 & 0.986 $\pm$ 0.001  &0.981 $\pm$ 0.003  & 0.892 $\pm$ 0.005 \\
JODIE & 0.828 ± 0.009  & 0.988 $\pm$ 0.002 & 0.991 $\pm$ 0.001 & 0.905 $\pm$ 0.009 \\
DyRep &  0.811 ± 0.002 & 0.857 $\pm$ 0.001  & 0.982 $\pm$ 0.001 & 0.891 ± 0.002 \\

TGAT   & 0.871 ± 0.002 & 0.992 $\pm$ 0.001 & 0.989 $\pm$ 0.001 & 0.785 $\pm$ 0.008 \\

TGN  & 0.913 ± 0.002 & 0.993 $\pm$ 0.001  & \underline{0.991} $\pm$ 0.001 & 0.921 $\pm$ 0.002 \\

GraphMixer  & 0.840 ± 0.002 & \underline{0.994} $\pm$ 0.001 & 0.990 $\pm$ 0.001 & 0.920 $\pm$ 0.005\\

DyGFormer & \underline{0.879} ± 0.006 & 0.993± 0.001 &0.990 $\pm$ 0.001 & \underline{0.946} $\pm$ 0.001\\
\hline
ScaDyG &\textbf{0.931} $\pm$ 0.009   &\textbf{0.996} $\pm$ 0.001 & \textbf{0.993} $\pm$ 0.002  & \textbf{0.955} $\pm$ 0.001 \\
\hline
\end{tabular}
}
\label{link_pred_auc}

\end{table}

\subsection{Additional analysis on results.}
\noindent
\label{baseline_results}
\textbf{Baseline performance.} Continuous-time-based methods have the highest complexity related to the number of edges due to their intricate designs for handling the neighborhood sequences of nodes associated with each edge. Since dynamic graphs typically have a significantly larger number of edges than nodes, this results in substantial time overhead.  In contrast, discrete-time-based methods exhibit lower complexity as their training time is primarily associated with node-related or simpler edge-related operations, thus exhibit faster training speed compared to continuous-time methods in experiments. Discrete-time-based and memory module-based methods exhibit node-related spatial complexity as they need to store the state of all nodes, leading to higher GPU memory usage compared to other methods in experiments. 
% Notably, JODIE is the second fastest method for link prediction,  but it performs much slower in node affinity prediction. 

\subsection{Relation to existing time encodings.}
\textcolor{black}{Existing time modeling techniques mainly fall into two categories: trigonometric-based encoding \cite{xu2020:inductive,cong2023:we} and exponential-based techniques \cite{zuo2018:embedding,lu2019:temporal,wen2022:trend}. The former vectorizes timestamps using functions with different parameters, and models time intervals through the dot product of time vectors.  While these techniques are expressive, they require learnable frequencies, and thus cannot be used in preprocessing. Exponential-based techniques, especially point processes-based methods, model the decay of historical interactions over time using a single exponential function. 
Our approach belongs to exponential-based techniques, but instead of using a single function, we use a combination of exponentials to enable dynamic time modeling. The various exponentials in a time encoding are set with predefined parameters, making it convenient for preprocessing. During the learning phase, the exponential functions are dynamically combined, offering greater representational power than a single exponential function.   }

% \begin{table}
%     \caption{Scalability performance on reddit dataset.}
% \begin{tabular}{cccccc}
% \hline
% Method &Pre & E-train  & E-eval &B-Mem & Param \\
% \hline
% EvolveGCN &- &  2 &  3& 901K\\
% ROLAND &- & Row 2, Col 2 & & & 834K \\

% DyRep &- & 90.56s  & 63.41s & 1600M      & 778K \\

% TGAT &- & 121.31 & 64.36s & 10052M & 802K\\

% TGN &- & 97.20s & 80.69s & 10752M & 935K \\

% GraphMixer &- & 45.35s & 59.36s & 10408M & 627K \\

% DyGFormer &- & 4863 s  &1144s &  10606M & 1072K\\
% \hline
% ScaDyG & 13.55s & 49s & s &  & 182K\\
% \hline
% \end{tabular}
% \end{table}

% \begin{table}
%     \caption{Scalability performance on tgbl-genre dataset.}
% \begin{tabular}{cccccc}
% \hline
% Method &Pre & E-train  & E-eval &B-Mem & Param \\
% \hline
% EvolveGCN &- & 59.54s & 68.25s & 9746M & 901K\\
% ROLAND &- &23.75s & 39.28s & 7524M& 401K \\

% DyRep &- & 90.56s  & 63.41s & 10556M      & 778K \\

% TGAT &- & 121.31 & 64.36s & 10052M & 802K\\

% TGN &- & 97.20s & 80.69s & 10752M & 935K \\

% GraphMixer &- & 45.35s & 59.36s & 10408M & 627K \\

% DyGFormer &- & 196.03 s  &41.02s &  10606M & 1072K\\
% \hline
% ScaDyG & 9.33s & 16.95s & 29.97s & 1266M & 158K\\
% \hline
% \end{tabular}
% \end{table}

\end{document}